\documentclass[10pt,twocolumn,letterpaper]{article}

\usepackage[pagenumbers]{cvpr} % To force page numbers, e.g. for an arXiv version

\usepackage{graphicx}
\makeatletter
\@namedef{ver@everyshi.sty}{}
\makeatother
\usepackage{epstopdf}
\usepackage{comment}
\usepackage{extarrows}
\usepackage{multirow}
\usepackage{amssymb}
\usepackage{threeparttable,booktabs}
\usepackage{floatrow}
\usepackage{diagbox}
\usepackage{array}
\usepackage{marginnote}
\usepackage{amsmath,amssymb} % define this before the line numbering.
\usepackage{color}
\usepackage{soul}
\usepackage{float}
\usepackage{subcaption}
\usepackage{wrapfig}
\usepackage{utfsym}
\usepackage{tabularx}
\floatstyle{plaintop}
\restylefloat{table}
\usepackage[accsupp]{axessibility}  % Improves PDF readability for those with disabilities.
\usepackage[symbol]{footmisc}

\usepackage{xcolor,colortbl}
\usepackage{pifont}
\usepackage{makecell}

\usepackage[pagebackref,breaklinks,colorlinks]{hyperref}

\usepackage[capitalize]{cleveref}
\Crefname{section}{Section}{Sections}
\Crefname{table}{Table}{Tables}
\crefname{table}{Tab.}{Tabs.}

\newcommand{\cmark}{\ding{51}}%
\newcommand{\xmark}{\ding{55}}%

\newfloatcommand{capbtabbox}{table}[][\FBwidth]

\newcommand{\tempri}[1]{%
\reflectbox{\scalebox{-1}[1]{\includegraphics[width=6pt]{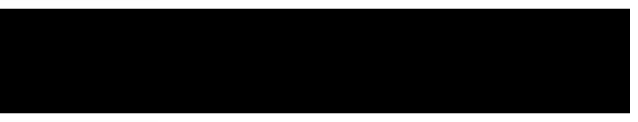}}}
}
\newcommand{\tempritransparent}[1]{%
\reflectbox{\scalebox{-1}[1]{\includegraphics[width=6pt]{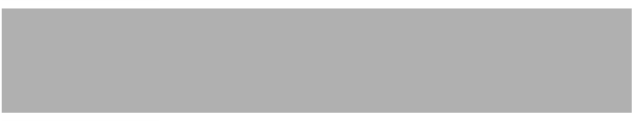}}}
}
\newcommand{\temprii}[1]{%
\reflectbox{\scalebox{-1}[1]{\includegraphics[width=6pt]{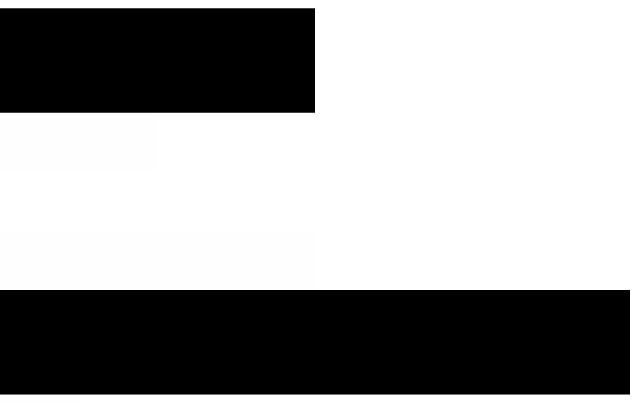}}}
}
\newcommand{\tempriitransparent}[1]{%
\reflectbox{\scalebox{-1}[1]{\includegraphics[width=6pt]{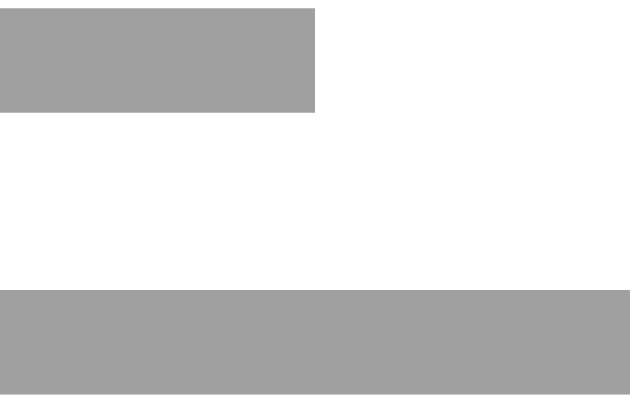}}}
}
\newcommand{\tempriii}[1]{%
\reflectbox{\scalebox{-1}[1]{\includegraphics[width=6pt]{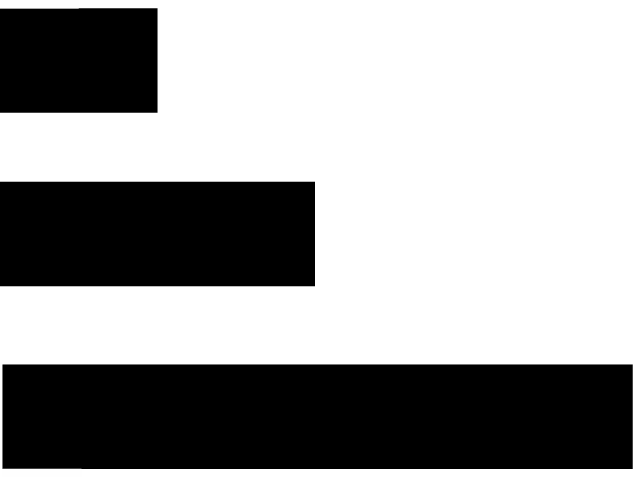}}}
}
\newcommand{\tempriiitransparent}[1]{%
\reflectbox{\scalebox{-1}[1]{\includegraphics[width=6pt]{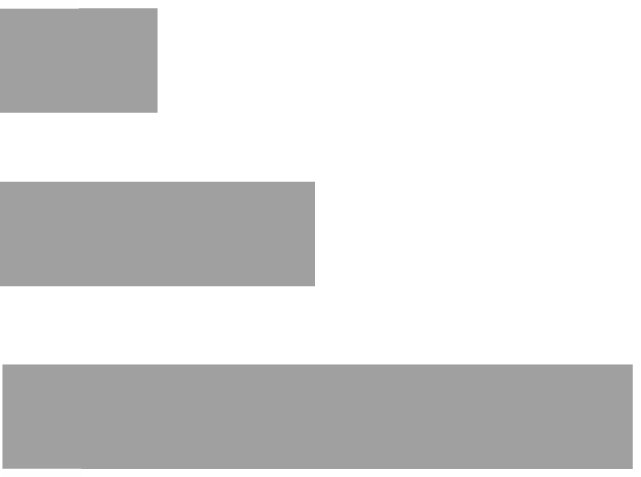}}}
}
\newcommand{\tempriv}[1]{%
\reflectbox{\scalebox{-1}[1]{\includegraphics[width=6pt]{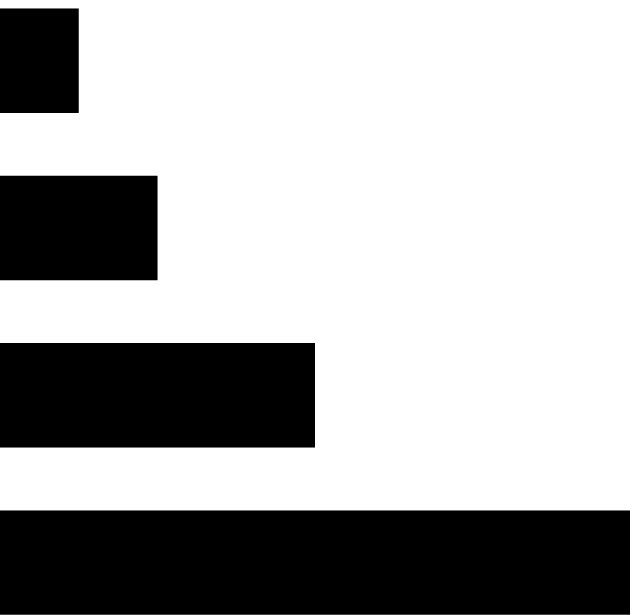}}}
}

\newcommand{\full}[1]{%
\reflectbox{\rotatebox[origin=c]{180}{\includegraphics[width=6pt]{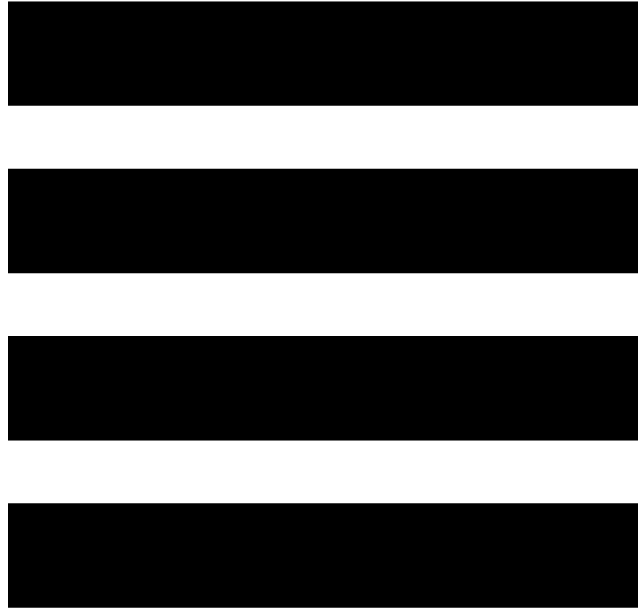}}}
}

\newcommand{\equal}[1]{%
\reflectbox{\rotatebox[origin=c]{180}{\includegraphics[width=6pt]{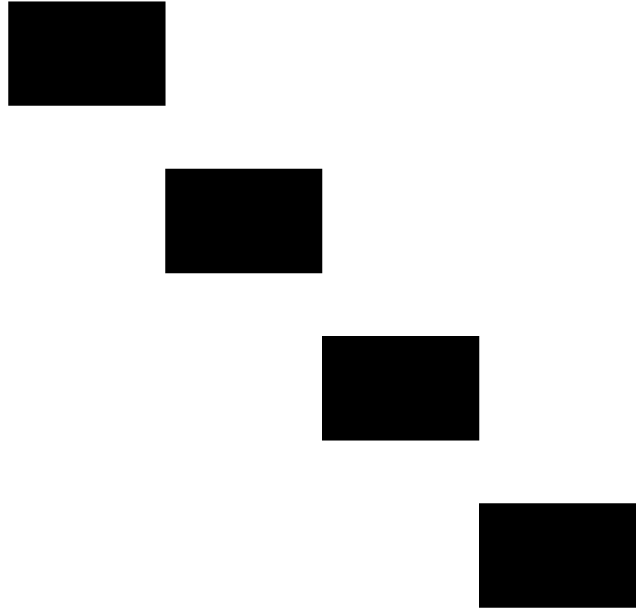}}}
}

\newcommand{\random}[1]{%
\reflectbox{\rotatebox[origin=c]{180}{\includegraphics[width=6pt]{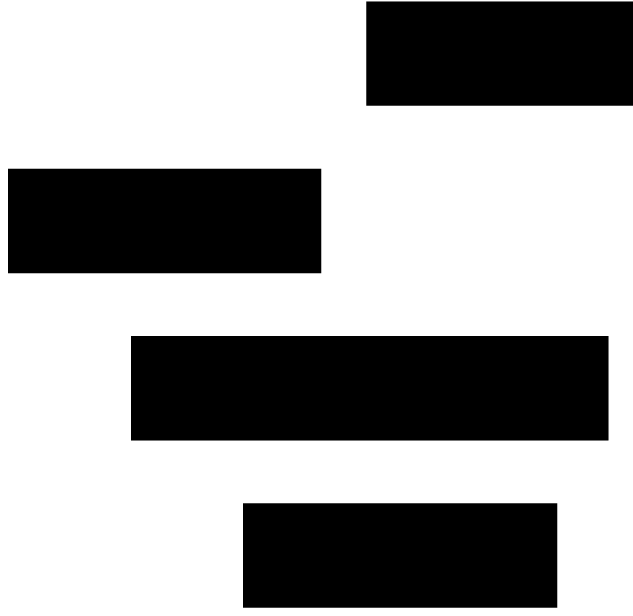}}}
}

\newcommand{\increasing}[1]{%
\reflectbox{\scalebox{-1}[1]{\includegraphics[width=6pt]{TemPr_eps/TemPr4_icon.eps}}}
}

\newcommand{\decreasing}[1]{%
\reflectbox{\scalebox{-1}[1]{\rotatebox[origin=c]{180}{\includegraphics[width=6pt]{TemPr_eps/TemPr4_icon.eps}}}}
}

\newcommand{\tempriva}[1]{%
\reflectbox{\scalebox{-1}[1]{\includegraphics[width=8pt]{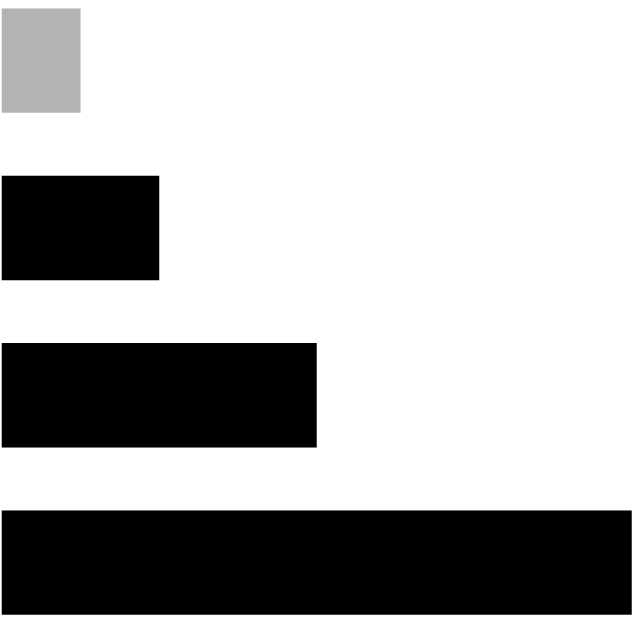}}}
}
\newcommand{\temprivb}[1]{%
\reflectbox{\scalebox{-1}[1]{\includegraphics[width=8pt]{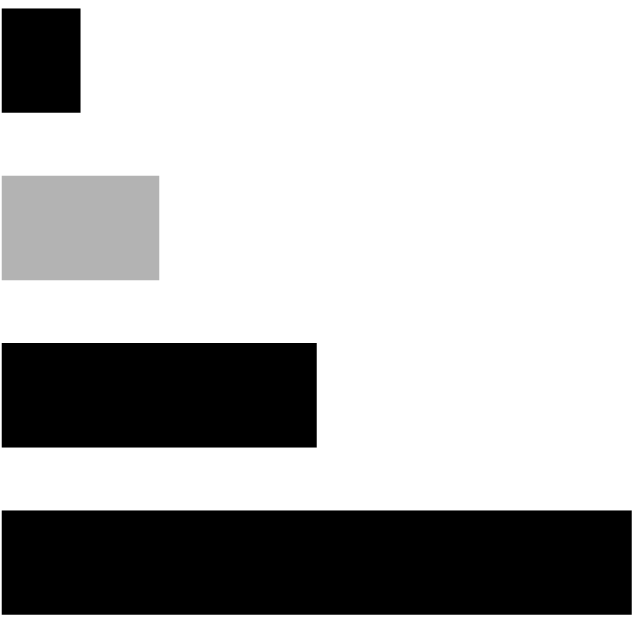}}}
}
\newcommand{\temprivc}[1]{%
\reflectbox{\scalebox{-1}[1]{\includegraphics[width=8pt]{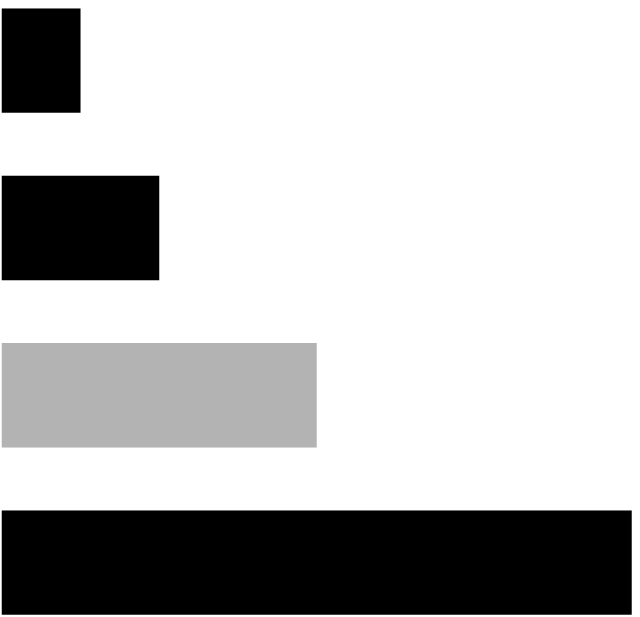}}}
}
\newcommand{\temprivd}[1]{%
\reflectbox{\scalebox{-1}[1]{\includegraphics[width=8pt]{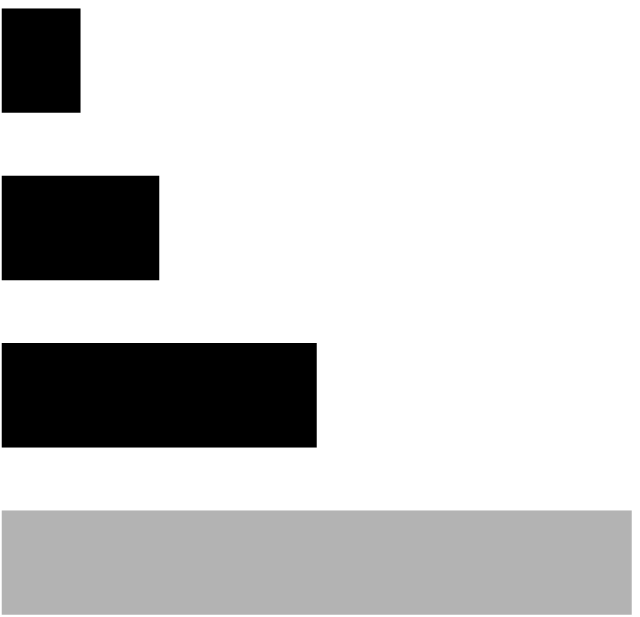}}}
}
\newcommand{\temprivab}[1]{%
\reflectbox{\scalebox{-1}[1]{\includegraphics[width=8pt]{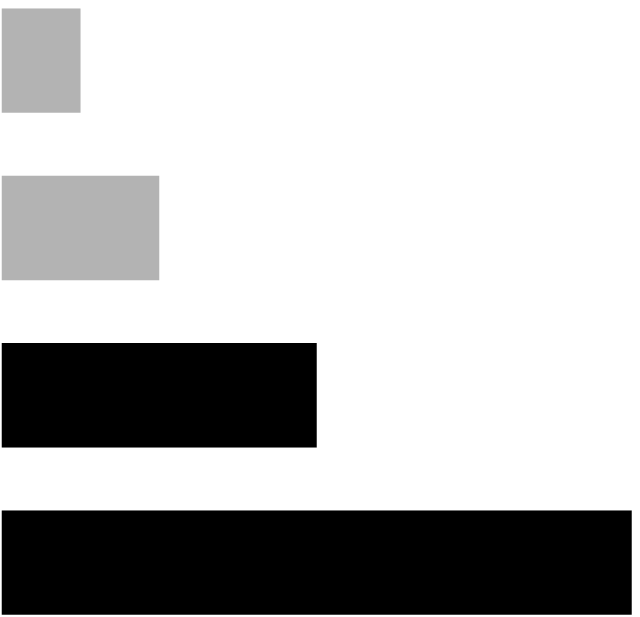}}}
}
\newcommand{\temprivac}[1]{%
\reflectbox{\scalebox{-1}[1]{\includegraphics[width=8pt]{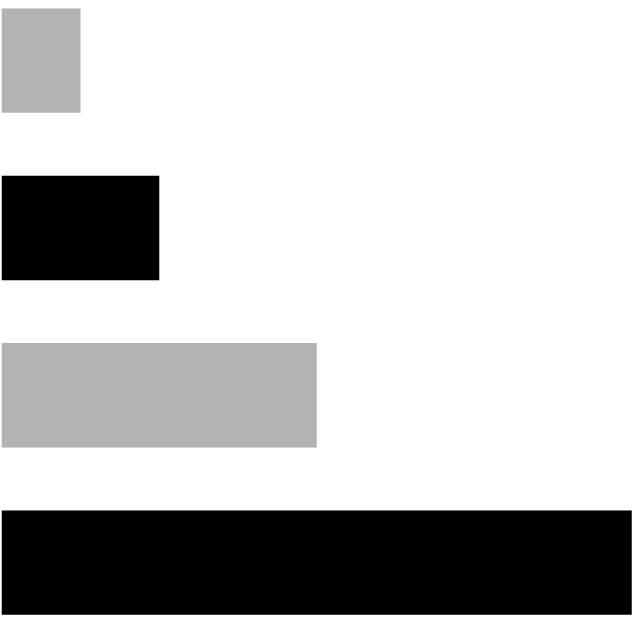}}}
}
\newcommand{\temprivad}[1]{%
\reflectbox{\scalebox{-1}[1]{\includegraphics[width=8pt]{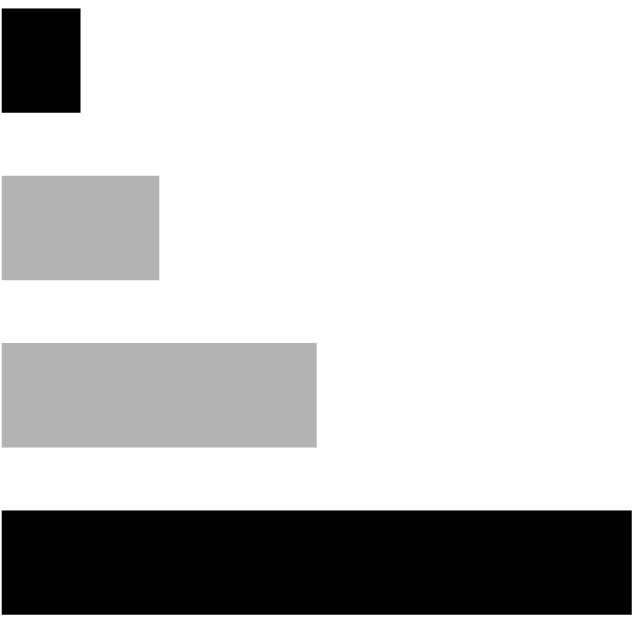}}}
}
\newcommand{\temprivbc}[1]{%
\reflectbox{\scalebox{-1}[1]{\includegraphics[width=8pt]{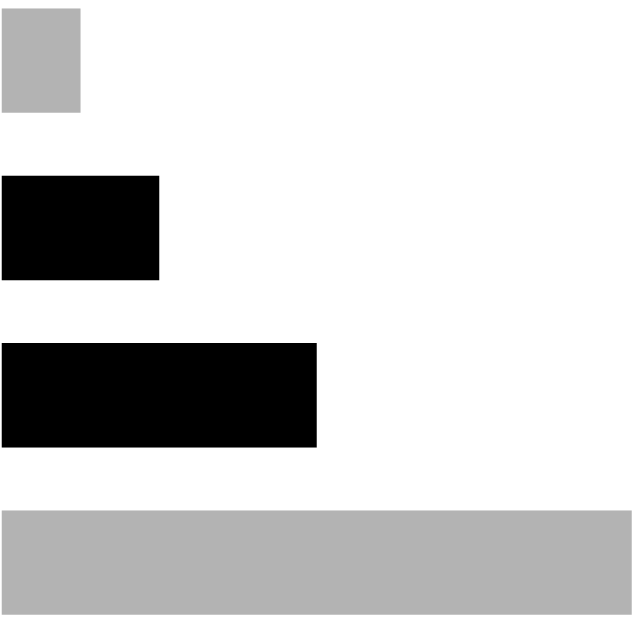}}}
}
\newcommand{\temprivbd}[1]{%
\reflectbox{\scalebox{-1}[1]{\includegraphics[width=8pt]{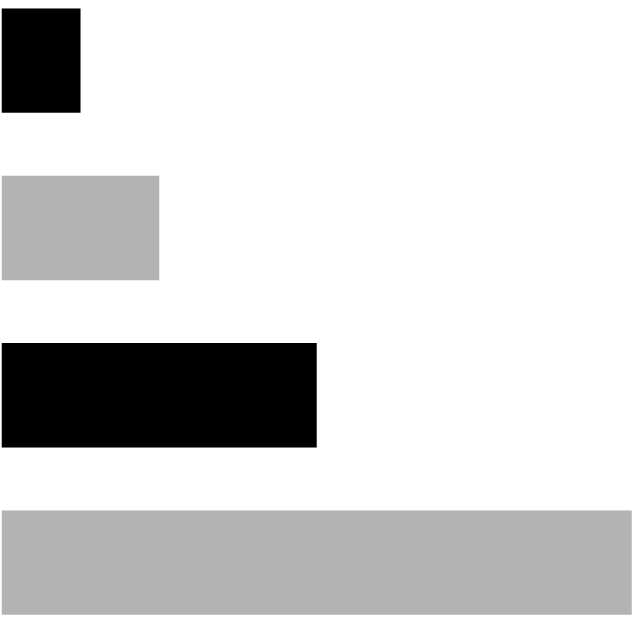}}}
}
\newcommand{\temprivcd}[1]{%
\reflectbox{\scalebox{-1}[1]{\includegraphics[width=8pt]{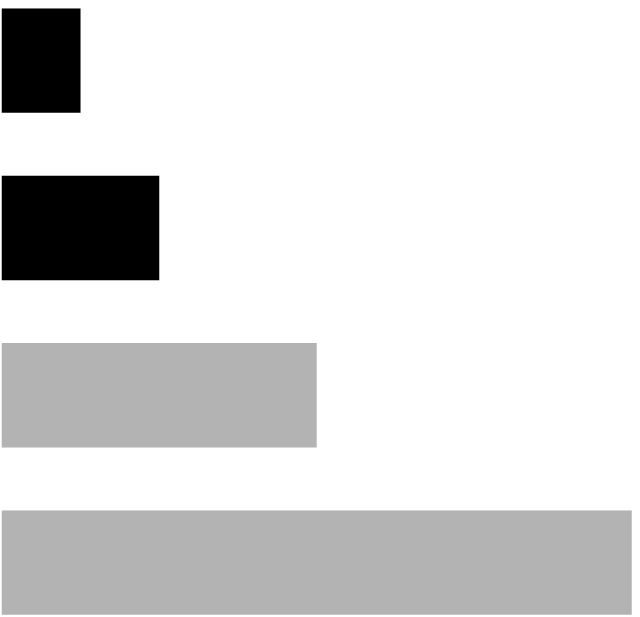}}}
}
\newcommand{\temprivabc}[1]{%
\reflectbox{\scalebox{-1}[1]{\includegraphics[width=8pt]{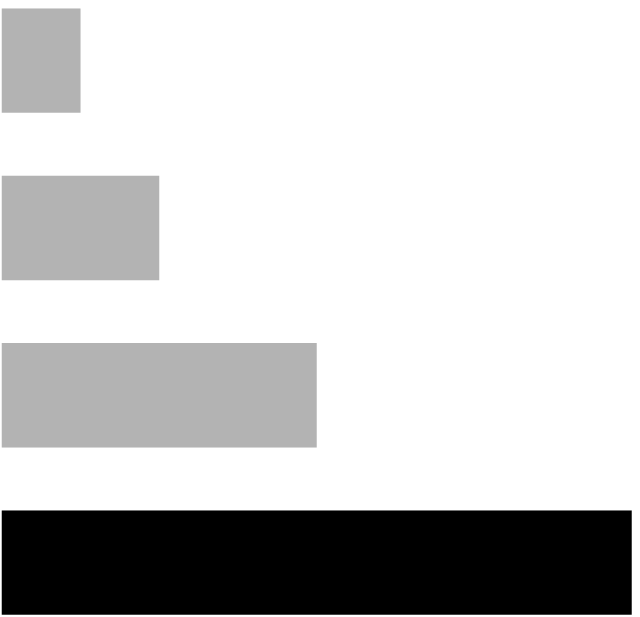}}}
}
\newcommand{\temprivabd}[1]{%
\reflectbox{\scalebox{-1}[1]{\includegraphics[width=8pt]{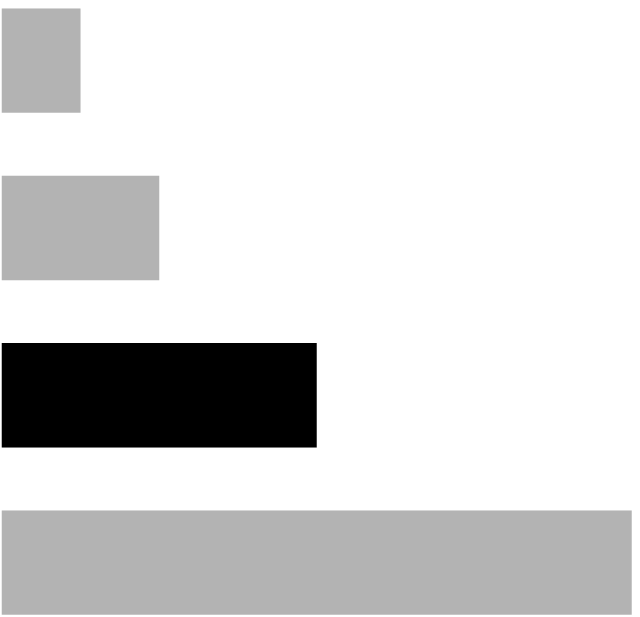}}}
}
\newcommand{\temprivacd}[1]{%
\reflectbox{\scalebox{-1}[1]{\includegraphics[width=8pt]{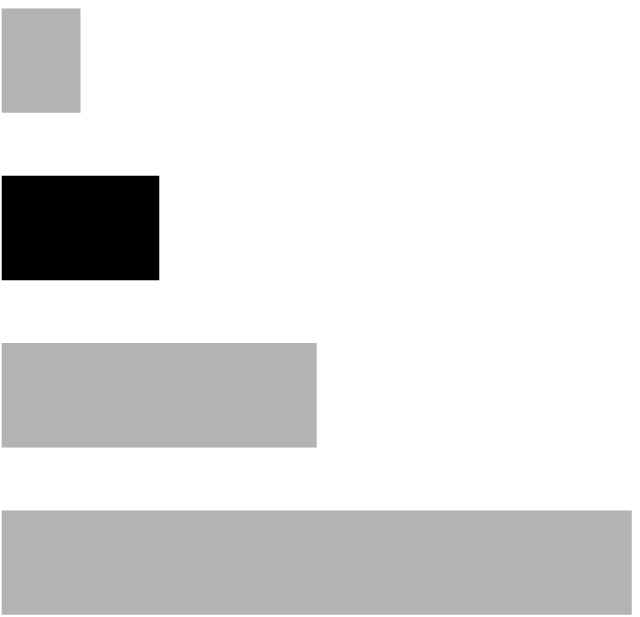}}}
}
\newcommand{\temprivbcd}[1]{%
\reflectbox{\scalebox{-1}[1]{\includegraphics[width=8pt]{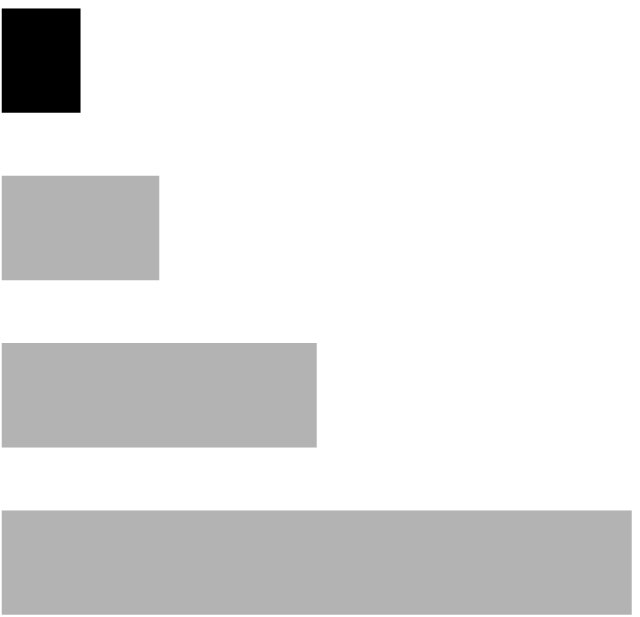}}}
}

\newcommand\tstrut{\rule{0pt}{2.4ex}}
\newcommand\bstrut{\rule[-1.0ex]{0pt}{0pt}}

\usepackage[labeled,resetlabels]{multibib}
\usepackage{xfp}
\newcommand\SupplementaryMaterials{%
  \xdef\presupfigures{\arabic{figure}}% save the current figure number
  \xdef\presupsections{\arabic{section}}% save the current section number
  \renewcommand\thefigure{S\fpeval{\arabic{figure}-\presupfigures}}
  \renewcommand\thesection{S\fpeval{\arabic{section}-\presupsections}}
  \renewcommand{\thetable}{S\arabic{table}}
  \renewcommand{\theequation}{S\arabic{equation}}
}

\definecolor{applegreen}{rgb}{0.0, 0.5, 0.0}
\definecolor{cadmiumred}{rgb}{0.89, 0.0, 0.13}
\definecolor{LightGrey}{rgb}{0.9,0.9,0.9}
\definecolor{basegray}{rgb}{0.88, 0.88, 0.88}

\definecolor{a1orange}{HTML}{FDC187}
\DeclareRobustCommand{\hla}[1]{{\sethlcolor{a1orange}\hl{#1}}}
\definecolor{a2red}{HTML}{EC8887}
\DeclareRobustCommand{\hlb}[1]{{\sethlcolor{a2red}\hl{#1}}}
\definecolor{a3green}{HTML}{96C8A8}
\DeclareRobustCommand{\hlc}[1]{{\sethlcolor{a3green}\hl{#1}}}
\definecolor{a4purple}{HTML}{B598C5}
\DeclareRobustCommand{\hld}[1]{{\sethlcolor{a4purple}\hl{#1}}}

\DeclareMathAlphabet\mathbfcal{OMS}{cmsy}{b}{n}

\def\cvprPaperID{331} % *** Enter the CVPR Paper ID here
\def\confName{CVPR}
\def\confYear{2023}

\begin{document}


\title{The Wisdom of Crowds: Temporal Progressive Attention \\ for Early Action Prediction}

\author{Alexandros Stergiou$^{1,2,*}$ \quad\quad Dima Damen$^{3}$ \\
\noindent $^1$Vrije University of Brussels, Belgium$\quad$ $^2$imec, Belgium $\quad$
$^3$University of Bristol, UK
%\\
%\tt\small{alexandros.stergiou@vub.be $\quad$ dima.damen@bristol.ac.uk}
}

\maketitle

%%%%%%%%% ABSTRACT
\begin{abstract}
\footnotetext[1]{Work carried out while A. Stergiou was at University of Bristol}
\setcounter{footnote}{1}
Early action prediction deals with inferring the ongoing action from partially-observed videos, typically at the outset of the video. We propose a bottleneck-based attention model that captures the evolution of the action, through progressive sampling over fine-to-coarse scales. Our proposed \textbf{Tem}poral \textbf{Pr}ogressive (TemPr) model is composed of multiple attention towers, one for each scale. The predicted action label is based on the collective agreement considering confidences of these towers. Extensive experiments over four video datasets showcase state-of-the-art performance on the task of Early Action Prediction across a range of encoder architectures.
We demonstrate the effectiveness and consistency of TemPr through detailed ablations.\footnote{Code is available at: \url{https://tinyurl.com/temprog}}

%This enables each tower to attend over progressively increasing temporal scales and to model relevant features, from fine-grained motions, to events over longer durations. 

%As only a portion of the video is observed, in many cases, typical recognition approaches are not suited for determining what action is being performed.
%may be insufficient to determine 

%\keywords{Early action prediction, attention, progressive sampling}
\end{abstract}

\section{Introduction}
\label{sec:intro}

% Overview of progress made
Early action prediction (EAP) is the task of inferring the action label corresponding to a given video, from only partially observing the start of that video.
Interest in EAP has increased in recent years due to both the ever-growing number of videos recorded and the requirement of processing them with minimal latency. Motivated by the advances in action recognition \cite{carreira2017quo,tran2018closer}, where the entire video is used to recognize the action label, recent EAP methods~\cite{cai2019action,fernando2021anticipating,kong2017deep,qin2017binary,wang2019progressive} distill the knowledge from these recognition models to learn from the observed segments. 
Despite promising results, the information that can be extracted from partial and full videos is inevitably different.
We instead focus on modeling the observed partial video better.

\begin{figure}[t]
\centering
\includegraphics[width=\linewidth]{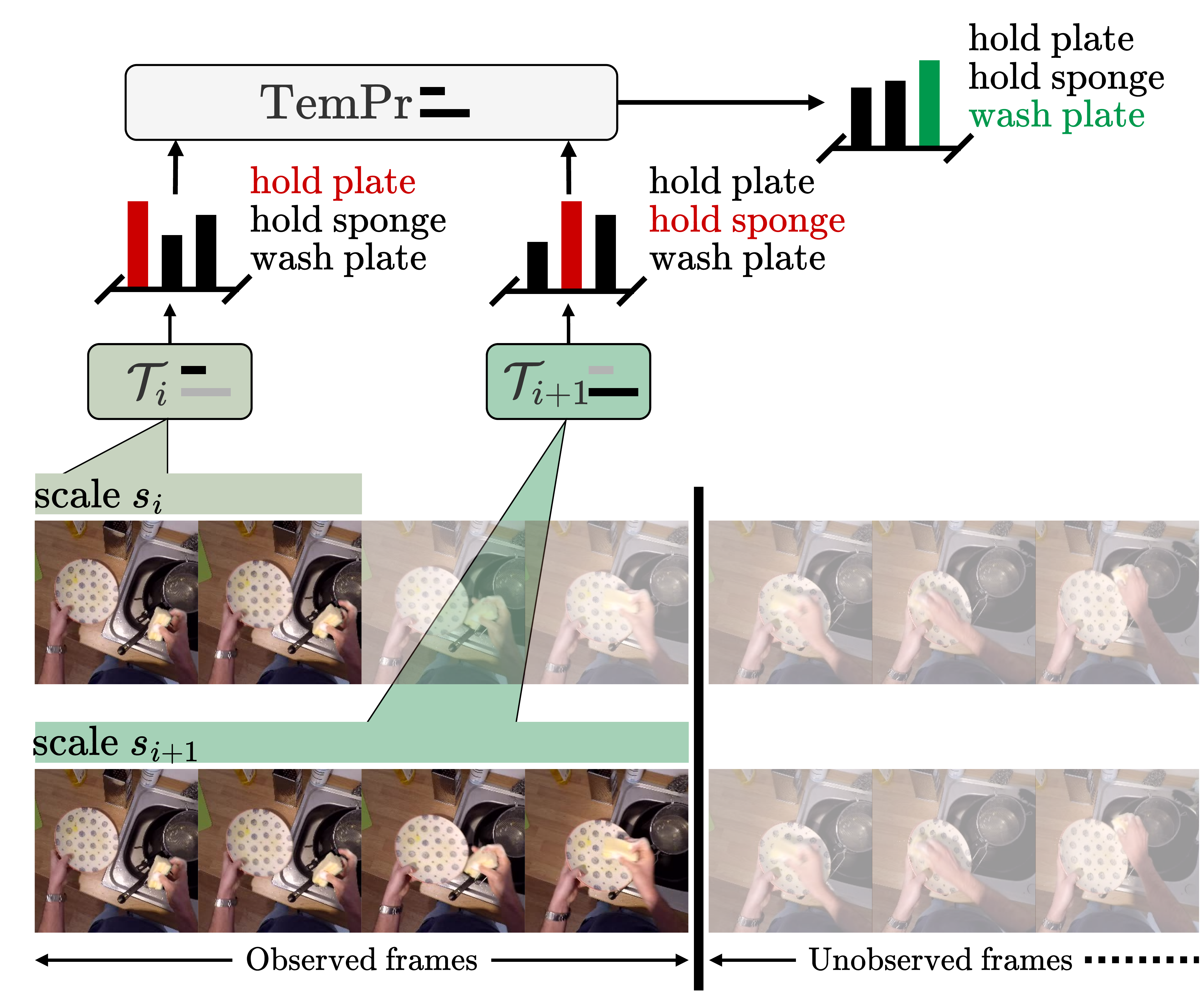}
\caption{\textbf{Early action prediction with TemPr} involves the use of multiple scales for extracting features over partially observed videos. Encoded spatio-temporal features are attended by distinct transformer towers ($\mathcal{T}$) at each scale.
We visualize two scales, where the fine scale $\mathcal{T}_i$ predicts `hold plate', and the coarse scale $\mathcal{T}_{i+1}$ predicts `hold sponge'.
Informative cues from both scales are combined for early prediction of the action `wash plate'.}
\vspace*{-8pt}
\label{fig:tempr_cover}
\end{figure}

% Task definition
Several neurophysiological studies \cite{fadiga1995motor,kohler2002hearing} have suggested that humans understand actions in a predictive and not reactive manner. This has resulted in the \textit{direct matching hypothesis} \cite{gallese1996action,rizzolatti1996premotor} where, actions are believed to be perceived through common patterns. 
Encountering any of these patterns prompts the expectation of specific action(s), even before the action is completed. Although the early prediction of actions is an inherent part of human cognition, the task remains challenging for computational modeling.

% Our approach
Motivated by the direct matching hypothesis, we propose a Temporally Progressive (TemPr) approach to modeling partially observed videos.
Inspired by multi-scale representations in images \cite{chen2019drop,zhang2021multi} and video \cite{hussein2019timeception,wu2022memvit}, we represent the observed video by a set of sub-sequences of temporally increasing lengths as in~\Cref{fig:tempr_cover}, which we refer to as scales.
TemPr uses distinct transformer towers over each video scale. These utilize a shared latent-bottleneck for cross-attention~\cite{jaegle2021perceiver,lee2019set}, followed by a stack of self-attention blocks to concurrently encode and aggregate the input. From tower outputs, a shared classifier produces label predictions for each scale. Labels are aggregated based on their collective similarity and individual confidences.

% Contributions
In summary, our contributions are as follows:
(i) We propose a progressive fine-to-coarse temporal sampling approach for EAP. (ii) We use transformer towers over sampled scales to capture discriminative representations and adaptively aggregate tower predictions, based on their confidence and collective agreement. (iii) We evaluate the effectiveness of our approach over four video datasets: UCF-101 \cite{soomro2012ucf101}, EPIC-KITCHENS~\cite{damen2022epic}, NTU-RGB \cite{shahroudy2016ntu} and Something-Something (sub-21 \& v2)~\cite{goyal2017something}, consistently outperforming prior work. 

% Related work
\section{Related Work}
\label{sec:related}

% Prediction - Recognition - anticipation
The task of EAP is related to but distinctly different from the tasks of action recognition and action anticipation. 
EAP predicts the \emph{ongoing} action label, partially observed.
In contrast, recognition assumes the \emph{completed} action has been fully observed, while anticipation forecasts potential \emph{upcoming} actions, seconds before the action starts.
We first review prior EAP approaches, before relating our method to those used for other video understanding tasks.

% Action prediction
\noindent \textbf{Early action prediction:}  Most of the early attempts have focused on the probabilistic modeling of partially observed videos \cite{cao2013recognize,lan2014hierarchical,li2014prediction,li2012modeling,ryoo2011human}. For example, Ryoo \etal ~\cite{ryoo2011human} used a bag-of-words approach to model feature distributions over multiple partially observed videos. Later approaches aimed to overcome errors where large appearance variations occur, by either sparse coded feature bases~\cite{cao2013recognize} or through a scoring function \cite{kong2015max,kong2014discriminative}, combining prior knowledge and the sequential order of frames. Lan \etal ~\cite{lan2014hierarchical} studied the representation of movements within the partially observed video, using a hierarchical structure. 

More recent methods \cite{cai2019action,fernando2021anticipating,hou2020confidence,hu2018early,kong2018action,wang2019progressive,wu2021anticipating,xu2019prediction,zhao2019spatiotemporal} have used learned-features. Specifically, knowledge distillation \cite{hinton2015distilling,park2019relational} has been used to transfer class knowledge from the complete videos to the corresponding partial videos. This was achieved using Long Short-Term Memory (LSTM) models~\cite{hu2018early,pang2019dbdnet,wang2019progressive} and teacher-student frameworks~\cite{cai2019action,fernando2021anticipating,wang2019progressive}. Other methods are based on recurrent architectures with additional memory cells \cite{kong2018action} for matching similar characteristics between the full and partial videos. Xu \etal ~\cite{xu2019prediction} proposed a conditional generative adversarial network
%\cite{mirza2014conditional} 
to generate feature representations for the entire video, from the partially observed video. Approaches have also focused on the propagation of residual features~\cite{zhao2019spatiotemporal} or exploration with graph convolutions through relation reasoning~\cite{wu2021spatial,wu2021anticipating}. Foo \etal ~ \cite{foo2022era} proposed specializing features during training into instance-specific and general features. Instance-specific features are learned from a subset of videos focusing on subtle cues, while general features are learned from the entire dataset.

In contrast, we hypothesize that it is more beneficial to represent the partial video progressively. Our method is based on sampling at varying-length scales from the observed video to understand the temporal progression of actions. We show that aggregating these predictors can lead to notable improvements in accuracy. To our knowledge, we are the first to study EAP in this progressive manner.

% Scales
\noindent \textbf{Multi-scale representations for other video understanding tasks}. The usage of scales, i.e. sequences of varying lengths or sampling at differing rates, is common in other video understanding task. For action recognition, video scales have been primarily used as a sampling method for either relational reasoning \cite{feichtenhofer2019slowfast,sermanet2017time,zhou2018temporal} or to select the most salient scale(s) as input to the network~\cite{meng2020ar,wu2019adaframe,zheng2020dynamic}. Xu \etal ~\cite{xu2021long} proposed the Long Short-Term Transformer, an encoder-decoder for relating current actions with their long-term context. In action anticipation, methods utilize different scales to combine features from video snippets and anticipate one or more upcoming actions \cite{furnari2020rolling,girdhar2021anticipative}. Different from these tasks, and based on the fact that informative parts of partially observed videos do not have fixed lengths, we propose to utilize progressive video scales, which capture fine-to-coarse representations making them more suitable for partially observed videos.

% Transformers
\noindent \textbf{Attention for video tasks}. Attention-based video methods~\cite{wang2018non,wu2019long} have initially been used as part of spatio-temporal CNNs \cite{carreira2017quo,tran2018closer}. The recent introduction of Vision Transformer \cite{dosovitskiy2020image} has inspired subsequent works on action recognition by either focusing on how spatio-temporal information can be processed \cite{arnab2021vivit,bertasius2021space} or architectural optimizations for spatio-temporal data~\cite{fan2021multiscale,liu2021video,ryoo2021tokenlearner,yan2022multiview,zhang2021vidtr}. Motivated by the recent advances of transformers for action recognition, we combine multiple transformer towers in TemPr.

\begin{figure*}[t]
\includegraphics[width=\linewidth]{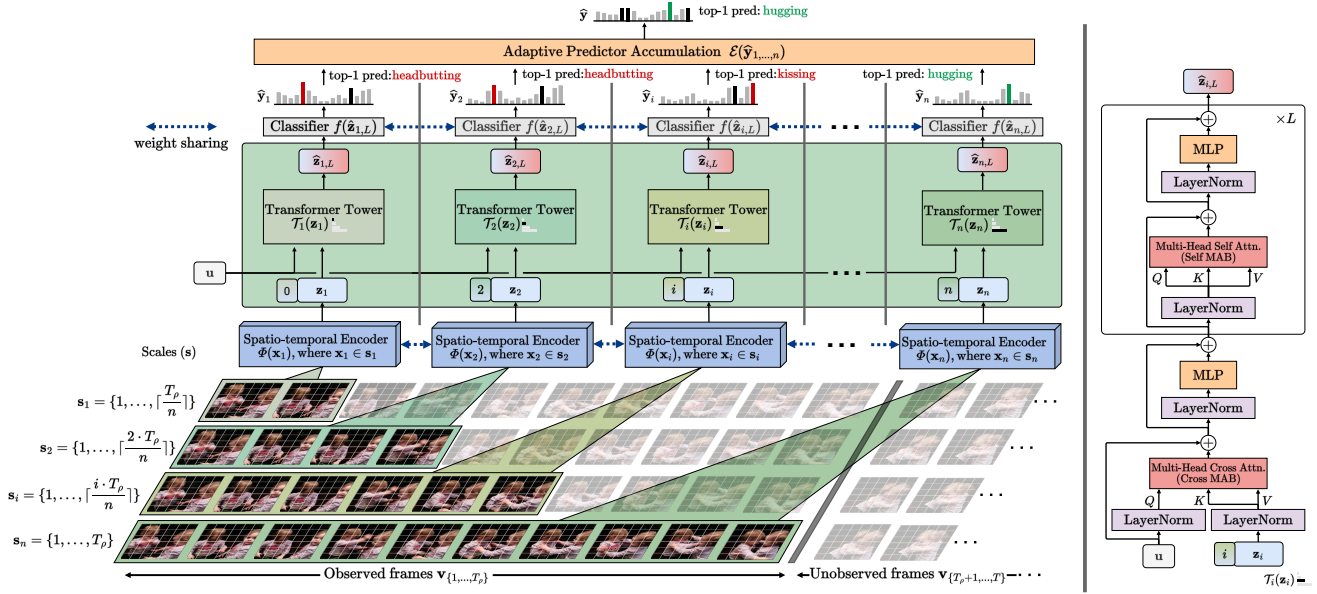}
\vspace*{-12pt}
\caption{(Left) \textbf{TemPr architecture}. Features are extracted over each input $\mathbf{x}_{i}$ sampled from video scale $\mathbf{s}_{i}$, and combined with scale and spatio-temporal positional encodings. The encoded features $\mathbf{z}_{i}$ are passed to attention towers $\mathcal{T}_i$ which output tensors $\widehat{\mathbf{z}}_{i,L}$ in the latent space. Shared-weight classifier $f(\cdot)$ is applied to every tower output to make per-scale predictions. These predictions are aggregated by aggregation function $\mathcal{E}(\cdot)$, for early action prediction over the observed frames. (Right) \textbf{Attention Tower}. Each utilizes pre-norm and a shared latent array $\mathbf{u}$ for the cross-attention block (Cross MAB). This is followed by a stack of $L$ self-attention blocks (Self MAB).} 
\label{fig:tempr_h}
\end{figure*}

\section{Our Approach}
\label{sec:method}

In this section, we overview our TemPr model (shown in \Cref{fig:tempr_h}).
We first introduce our prime contribution of progressive scales for sampling from the observed video in \Cref{sec:method:sub::sampling}.
Each scale corresponds to an attention tower, which captures the progression of the action, and predicts the ongoing action, as explained in \Cref{sec:method:sub::attention}. Multiple scales/towers are then combined for a final prediction by an aggregation function, detailed in \Cref{sec:method:sub::ensemble}.  
% Overview and section structure

\subsection{EAP: Problem Definition}
\label{sec:method:sub::problem_def}

% Definition
We follow the standard definition of the EAP task from recent works \cite{cai2019action,wu2021anticipating,xu2019prediction,zhao2019spatiotemporal}. We denote the full video with $T$ frames as $\mathbf{v}_{\{1,...,T\}}$. 
We define the \emph{observation ratio} $0 \! < \! \rho \! < \! 1$ as the proportion of frames observed.
EAP assumes $0 < \rho$, i.e. at least one frame of the video depicting the action has been observed, and $\rho < 1$, i.e. part of the video remains unobserved.
Accordingly, $T_\rho = \lceil \rho \cdot T \rceil$ is the number of observed frames. In EAP, the prediction of the ongoing action label $y$ conveyed in the full video $\mathbf{v}_{\{1,...,T\}}$ is attempted from only the observed $T_{\rho}$ frames.

\subsection{Progressive Video Scales}
\label{sec:method:sub::sampling}

% Motivation
Given the partial observation of the action, we speculate that the sampling strategy is critical for capturing distinctive representations of the ongoing action. This is different from the sampling typically utilized in action recognition, where the video is uniformly split into equally-sized segments \cite{wang2016temporal}.
Equal-sized segments, in partially observed videos, can miss the discriminative action pattern when this pattern spans across segments. We thus propose to sample at multiple scales within the observed video, which we refer to as \emph{progressive} sampling.

% Scale selection
Given the partially observed video of $T_{\rho}$ frames, we examine the ongoing action over $n$ scales $\mathbf{s}_{\{1,..,n\}}$. Each scale $\mathbf{s}_{i}$ has a larger temporal extent to sample from than $\mathbf{s}_{i-1}$.
We represent each scale $\mathbf{s}_{i}$ as:
\begin{equation}
\label{eq:scale}
    \mathbf{s}_{i} \! = \! \{1,...,T_{\mathbf{s}_{i}} \} \;; \enspace T_{\mathbf{s}_{i}} \! = \! \lceil \frac{i}{n} \cdot T_{\rho} \rceil \; \forall \; i \in \mathbf{N} \! = \! \{1,...,n\}
\end{equation}
%from which, we sample $F$ frames randomly. 
Over each scale, we sample $F$ frames randomly to capture a progressive fine-to-coarse representation. Considering the variable input length per scale, sampling a fixed number of frames $F$, is required to standardize the encoder inputs.

\subsection{Temporal Progressive Attention Towers}
\label{sec:method:sub::attention}

% Overview
We use a shared encoder $\Phi(\cdot)$ to extract features from the sampled frames, over the progressive scales.
Corresponding to each scale $\mathbf{s}_{i}$, we define input volume~$\mathbf{x}_{i}$ of size $3 \! \times \!F \! \times \! H \! \times \! W$, with $F$ temporally ordered frames, $H$ height and $W$~width. We thus define ${\mathbf{z}_i = \Phi(\mathbf{x}_i)}$ to be the per-scale, multi-dimensional spatio-temporal encoded feature volume, of size $C \times t \times h  \times w$.
Given the scales' spatio-temporal features $\mathbf{z}_{1},...,\mathbf{z}_{n}$ , we reshape these to $C \! \times \! (thw)$, and concatenate Fourier Positional Embeddings (PE) of size $n \! \times \! (thw)$ to encode each scale and space-time position.
Features $\textbf{z}_{i}$ form the input to attention tower $\mathcal{T}_{i}$.

% Tower notations 
We attend each scale's features using tower $\mathcal{T}_{i}$, so that $\widehat{\mathbf{z}}_{i} \! = \! \mathcal{T}_{i}(\mathbf{z}_{i})$,
where $\widehat{\mathbf{z}}_{i}$ is the feature volume after attending input volume $\mathbf{z}_{i}$ over the transformer blocks. 
Motivated by the recent architectural approaches for dealing with the quadratic scaling of complexity in transformers \cite{jaegle2021perceiver,lee2019set}, each tower uses two attention components consisting of one cross-attention bottleneck block and a stack of self-attention blocks as shown in \Cref{fig:tempr_h} (right). 
Towers are indexed by $i \in \mathbf{N}$ and attention blocks, per tower, are indexed by ${j \in \{0,..,L\}}$. We describe these components next.

% Cross-attention
\noindent
\textbf{Cross Multi-Head Attention Block} (Cross MAB), employs a latent array $\mathbf{u}$ of $C \times d$ size ($d \! \ll \! thw$). This latent array alongside $\mathbf{z}_i$ are used to create the asymmetric query-key-value (\textbf{QKV}) attention function in which $\mathbf{Q} \! \in \! \mathbb{R}^{C \times d}$, $\mathbf{K} \! \in \! \mathbb{R}^{C \times (thw)}$, $\mathbf{V} \! \in \! \mathbb{R}^{C \times (thw)}$. The Cross MAB block consists of Multi-Head Cross Attention (MCA), Layer Normalization (LN), and Multilayer Perceptron (MLP) modules:
\begin{equation}
\begin{aligned}
\label{eq:cross_att}
    & \widehat{\mathbf{z}}_{i,0} = MLP(LN(\mathbf{h}_{i,0})) + \mathbf{h}_{i,0}, \, \text{where}\\
    & \mathbf{h}_{i,0} = MCA(LN(\mathbf{u}),LN(\mathbf{z}_{i})) + \mathbf{u} \, \;\forall \, i \in \mathbf{N}
\end{aligned}
\end{equation}
in which, the MCA computes the dot-product asymmetric attention of tensors $\mathbf{u}$ and $\mathbf{z}_{i}$. %, where $\mathbf{u}$ forms the query and $\mathbf{z}_{i}$ forms the keys and values pairs. 

By exploiting the Cross MAB~\cite{jaegle2021perceiver} bottleneck, the transformer towers are significantly more efficient than a deep stack of self-attention blocks. The use of a parameterizable size latent vector can benefit the creation of performance-balanced models, minimizing feature redundancies.

% Self attention
\noindent
\textbf{Stacked Self-Attention Blocks} (Self MAB), correspond to a stack of $L$ transformer blocks \cite{dosovitskiy2020image}, symmetrically attending to tensors ${\widehat{\mathbf{z}}_{i,j} \, \forall j \! \in \{0,...,L-1\}}$. Including Multi-Head Self Attention (MSA), the block is denoted as:
\begin{equation}
\begin{aligned}
\label{eq:self_attn}
    & \widehat{\mathbf{z}}_{i,j} = MLP(LN(\mathbf{h}_{i,j})) + \mathbf{h}_{i,j}, \, \text{where}\\
    & \textbf{h}_{i,j} = MSA(LN(\widehat{\mathbf{z}}_{i,j\text{-}1})) + \widehat{\mathbf{z}}_{i,j\text{-}1} \, \forall \, i \in \mathbf{N}, j \in \{1,...,L\}
\end{aligned}
\end{equation}

% Predictors
\noindent
\textbf{Attention tower predictors}. Towers additionally include a linear classifier $\widehat{\mathbf{y}}_i = f(\widehat{\mathbf{z}}_{i,L})$ that maps the output $\widehat{\mathbf{z}}_{i,L}$ to $\widehat{\mathbf{y}}_i$ class predictions. 
As features $\widehat{\mathbf{z}}_{i,L}$ are bound to scale $\textbf{s}_{i}$, towers cannot relate features across scales, which limits their modeling capabilities. 
We thus share classifier weights across scales to establish a joint feature space.

Predictions from the $n$ attention towers are thus obtained. We describe our proposed aggregation approach next.

\subsection{Aggregation Function for EAP}
\label{sec:method:sub::ensemble}

We wish to accumulate class predictions from the individual fine-to-coarse scales into an overall EAP for the observed $T_\rho$ frames. 

We introduce an aggregation function $\mathcal{E}(\widehat{\mathbf{y}}_{1,...,n})$ for accumulating tower predictions. The function is formulated based on the agreement between predictions and the individual towers' confidence in the produced prediction.

% Agreement
\noindent
\textbf{Predictor agreement}. We trust that predictions with a high degree of resemblance, in terms of their class probability distribution, can reduce the uncertainty of individual predictors. We utilize Exponential Inverse Coefficient Weighting~($e$ICW)~\cite{stergiou2021adapool} for the weighted aggregation of probabilities $\widehat{\mathbf{y}}_{i}$ per scale, based on their similarity to the mean probability distribution $\overline{\widehat{\mathbf{y}}}$: 
\begin{equation}
    \underset{\scriptstyle eICW}{\mathcal{E}} (\widehat{\mathbf{y}}_{i},\overline{\widehat{\mathbf{y}}}) = \frac{ e^{DSC(\widehat{\mathbf{y}}_{i},\overline{\widehat{\mathbf{y}}})^{-1}} }{\underset{k \in \mathcal{\mathbf{N}}}{\sum}e^{DSC(\widehat{\mathbf{y}}_{k},\overline{\widehat{\mathbf{y}}})^{-1}}} \cdot \widehat{\mathbf{y}}_{i}
\end{equation}
in which $DSC(\cdot)$ is the Dice-S\o rensen coefficient \cite{dice1945measures} between class probabilities~$\widehat{\mathbf{y}}_{i}$ and mean probabilities $\overline{\widehat{\mathbf{y}}}$. 

% Confidence
\noindent
\textbf{Predictor confidence}. Aggregation is performed based on the sharpness of the probability distribution. We calculate the exponential maximum (i.e. softmax) across all predictions. Predictions with high class probability for a single or a small set of classes are weighted higher: 
\begin{equation}
    \underset{\scriptstyle eM}{\mathcal{E}}(\widehat{\mathbf{y}}_{i}) = \frac{ e^{\widehat{\mathbf{y}}_{i}}}{\underset{k \in \mathbf{N}}{\sum}e^{\widehat{\mathbf{y}}_{k}}}  \cdot \widehat{\mathbf{y}}_{i}
\end{equation}

A combination of the two strategies is used for the final adaptive predictor aggregation function $\mathcal{E}(\widehat{\mathbf{y}}_{1,...,n})$. As in~\cite{stergiou2021adapool}, we use a parameter  ${0 \! \leq \! \beta \! \leq \! 1}$, which we learn during training, to determine the proportion of each method: 
\begin{equation}
    \mathcal{E}(\widehat{\mathbf{y}}_{1,...,n})) = \underset{i \in \mathbf{N}}{\sum} \; \beta \cdot \!\!\! \underset{\scriptstyle eICW}{\mathcal{E}}(\widehat{\mathbf{y}}_{i},\overline{\widehat{\mathbf{y}}}) + (1-\beta) \cdot \underset{\scriptstyle eM}{\mathcal{E}}(\widehat{\mathbf{y}}_{i})
\end{equation}
We refer to this aggregation function as our proposed \emph{adaptive} aggregation function for attention tower predictions. 

During training, we use the adaptive probability distribution from $\mathcal{E}(\widehat{\mathbf{y}}_{1,...,n})$ to calculate the divergence from the target one-hot categorical distribution for class vector $\mathbf{y}$.
In inference, the $\arg\max$ class is used as the EAP label.

In summary, our proposed method combines progressive scales of the observed video, individual attention towers with shared classifier weights, and an aggregation function that backpropagates through all individual attention towers. We evaluate our method next. 

\section{Experiments}
\label{sec:experiments}

% Overview
The datasets used, alongside implementation and training scheme details, are explained in \Cref{sec:experiments:sub::settings}. We include state-of-the-art model comparisons in \Cref{sec:experiments:sub::results} followed by ablation studies in \Cref{sec:experiments:sub::ablation}.

\begin{table*}[t]
\caption{\textbf{Top-1 accuracies (\%) of action prediction methods on UCF-101 over different observation ratios ($\mathbf{\rho}$)}. Methods are grouped w.r.t. the backbone used. We report \textbf{TemPr} results on 5 backbones. The best results per $\mathbf{\rho}$ are in \textbf{bold} and second best are \underline{underlined}.} 
\label{tab:ucf101_sota}
\centering
\resizebox{.92\textwidth}{!}{%
\begin{tabular}{l| l l | c c c c c c c c c }
\hline
\multicolumn{1}{c|}{\multirow{2}{*}{Method}} &
\multicolumn{1}{c}{\multirow{2}{*}{Backbone}} &
\multicolumn{1}{c|}{\multirow{2}{*}{dim}} &
\multicolumn{9}{c}{Observation ratios ($\rho$)} \tstrut \\
&
&
&
0.1 &
0.2 &
0.3 &
0.4 &
0.5 &
0.6 &
0.7 &
0.8 &
0.9  \tstrut \bstrut \\
\hline 

RGN-KF \cite{zhao2019spatiotemporal} &
\multirow{3}{*}{Inception \cite{szegedy2015going}} &
\multirow{3}{*}{2D} &
83.3 & 
85.2 &
87.8 &
90.6 &
91.5 &
92.3 &
92.0 &
93.0 &
92.9 \tstrut \\[.1em]

GGNN \cite{wu2021anticipating} &
&
&
82.4 & 
85.6 &
89.0 &
- &
91.3 &
- &
92.4 &
- &
93.0 \\[.1em]

TS (2$\! \times \!$L) \cite{wang2019progressive} &
&
&
83.3 & 
87.1 &
88.9 &
89.8 &
90.9 &
91.0 &
91.3 &
91.2 &
91.3 \bstrut \\
\hline

AAPNet \cite{kong2018adversarial} &
C3D \cite{tran2015learning} &
3D &
59.9 & 
80.4 &
86.8 &
86.5 &
86.9 &
88.3 &
88.3 &
89.9 &
90.9 \tstrut \bstrut \\
\hline

MSSC \cite{cao2013recognize} &
\multirow{8}{*}{ResNet-18} &
\multirow{5}{*}{2D \cite{he2016deep}} &
34.1 & 
53.8 &
58.3 &
57.6 &
62.6 &
61.9 &
63.5 &
64.3 &
62.7 \tstrut \\[.1em]

MTSSVM \cite{kong2014discriminative} &
&
&
40.1 & 
72.8 &
80.0 &
82.2 &
82.4 &
83.2 &
83.4 &
83.6 &
83.7 \\[.1em]

DeepSCN \cite{kong2017deep} &
&
&
45.0 & 
77.7 &
83.0 &
85.4 &
85.8 &
86.7 &
87.1 &
87.4 &
87.5 \\[.1em]

mem-LSTM \cite{kong2018action} &
&
&
51.0 & 
81.0 &
85.7 &
85.8 &
88.4 &
88.6 &
89.1 &
89.4 &
89.7 \\[.1em]

MSRNN \cite{hu2018early} &
&
&
68.0 & 
87.2 &
88.2 &
88.8 &
89.2 &
89.7 &
89.9 &
90.3 &
90.4 \\ [.1em]

GGNN \cite{wu2021anticipating} &
&
&
75.9 & 
81.7 &
87.8 &
- &
88.7 &
- &
89.4 &
- &
90.2 \\ [.1em]

\textbf{TemPr} \tempriv \ \textbf{(ours)} &
&
3D \cite{hara2018can} &
84.3 & 
90.2 &
90.4 &
90.9 &
91.2 &
91.8 &
92.1 &
92.3 &
92.4 \bstrut \\
\hline

AA-GAN \cite{gammulle2019predicting} &
\multirow{4}{*}{ResNet-50} &
\multirow{2}{*}{2D \cite{he2016deep}} &
- & 
84.2 &
- &
- &
85.6 &
- &
- &
- &
- \tstrut \\[.1em]

GGNN \cite{wu2021anticipating} & 
&
&
84.1 & 
88.5 &
89.8 &
- &
90.9 &
- &
91.4 &
- &
91.8 \\[.1em]

TS+JVS+JCC+JFIP \cite{fernando2021anticipating} &
&
&
- & 
85.8 &
- &
- &
- &
- &
- &
- &
- \\ [.1em]

\textbf{TemPr} \tempriv \ \textbf{(ours)}  &
&
3D \cite{hara2018can} &
84.8 & 
90.5 &
91.2 &
91.8 &
91.9 &
92.2 &
92.3 &
92.4 &
92.6 \bstrut \\
\hline

DBDNet \cite{pang2019dbdnet} &
\multirow{4}{*}{ResNeXt101 \cite{hara2018can}} &
\multirow{4}{*}{3D} &
82.7 & 
86.6 &
88.3 &
89.7 &
90.6 &
91.2 &
91.7 &
91.9 &
92.0 \tstrut \\[.1em]

IGGNN \cite{wu2021spatial} &
&
&
80.2 &
- &
89.8 &
- &
92.9 &
- &
94.1 &
- &
94.4 \bstrut \\

ERA \cite{foo2022era} & 
&
&
\textbf{89.1} & 
- & 
92.4 & 
- & 
94.3 & 
- & 
\underline{95.4} & 
- & 
95.7 \\ [.1em]

\textbf{TemPr} \tempriv \ \textbf{(ours)} &
&
&
85.7 & 
91.4 & 
92.1 & 
92.7 & 
93.5 & 
\underline{93.9} & 
94.4 & 
94.6 & 
94.9 \bstrut \\
\hline

\textbf{TemPr} \tempriv \ \textbf{(ours)} &
X3D$_{M}$ \cite{feichtenhofer2020x3d} &
3D &
87.9 & 
\underline{93.4} & 
\underline{94.5} & 
\underline{94.8} & 
\underline{95.1} & 
\textbf{95.2} & 
\textbf{95.6} & 
\underline{96.4} & 
\textbf{96.3} \tstrut \bstrut \\

\hline

\textbf{TemPr} \tempriv \ \textbf{(ours)} &
MoViNet-A4 \cite{kondratyuk2021movinets} &
3D &
\underline{88.6} & 
\textbf{93.5} & 
\textbf{94.9} & 
\textbf{94.9} & 
\textbf{95.4} & 
\textbf{95.2} & 
95.3 & 
\textbf{96.6} & 
\underline{96.2} \tstrut \\

\textcolor{gray}{TemPr} \tempriiitransparent \  &
\multirow{3}{*}{\textcolor{gray}{MoViNet-A4}} &
\multirow{3}{*}{\textcolor{gray}{3D}} &
\textcolor{gray}{87.3} & 
\textcolor{gray}{93.1} & 
\textcolor{gray}{94.9} & 
\textcolor{gray}{94.6} & 
\textcolor{gray}{95.2} & 
\textcolor{gray}{94.9} & 
\textcolor{gray}{94.6} & 
\textcolor{gray}{95.1} & 
\textcolor{gray}{95.0} \\[.1em]

\textcolor{gray}{TemPr} \tempriitransparent \  &
& &
\textcolor{gray}{85.6} & 
\textcolor{gray}{92.9} & 
\textcolor{gray}{93.6} & 
\textcolor{gray}{94.5} & 
\textcolor{gray}{94.4} & 
\textcolor{gray}{94.2} & 
\textcolor{gray}{94.2} & 
\textcolor{gray}{94.6} & 
\textcolor{gray}{94.8} \\[.1em]

\textcolor{gray}{TemPr} \tempritransparent \  & & &
\textcolor{gray}{85.2} & 
\textcolor{gray}{92.1} & 
\textcolor{gray}{92.5} & 
\textcolor{gray}{92.9} & 
\textcolor{gray}{93.3} & 
\textcolor{gray}{93.7} & 
\textcolor{gray}{93.5} & 
\textcolor{gray}{93.8} & 
\textcolor{gray}{93.7} \\[.1em]

\end{tabular}
}
\hfill
\end{table*}

\subsection{Datasets and Implementation Details}
\label{sec:experiments:sub::settings}

% Overview
\noindent
\textbf{Datasets} We report our method's performance over a diverse set of video datasets previously used for EAP. 
{\textit{UCF-101}~\cite{soomro2012ucf101}} consists of 101 classes and 13K videos depicting various types of actions such as human-object interactions, human-human interactions, playing musical instruments, and sports. \textit{Something-Something} (SSv1/SSsub21/SSv2)~\cite{goyal2017something} is a collection of 100K (SSv1) \& 220K (SSv2) videos of 174 fine-grained human-object action and interaction categories. The v1 of the dataset also includes a 21-action categories subset (SSsub21) of 11K videos used previously by \cite{wu2021spatial,wu2021anticipating} for EAP. We report on this subset, for direct comparisons and v2 for large-scale benchmarking. \textit{EPIC-KITCHENS-100} (EK-100)~\cite{damen2022epic} contains unscripted egocentric actions and activities across 45 kitchen environments. Labels are composed of 97~verb classes, 300 noun classes, and 4025 action classes of combined nouns and verbs. We also use the RGB-only version of \textit{NTU RGB+D} \cite{shahroudy2016ntu}, as in \cite{kong2017deep,li2014prediction}, containing 60 action classes and 57K videos of daily human actions.

Previous EAP works~\cite{cao2013recognize,cai2019action,kong2018action,kong2017deep,sadegh2017encouraging,wang2019progressive,wu2021spatial,wu2021anticipating} have evaluated their performance over smaller datasets ($<100K$ videos) that are only partially indicative of the approaches' generalizability. We thus set new EAP baselines by evaluating on two large-scale datasets: the temporally challenging SSv2 as well as EK-100.

% Architectural and parameter-based settings
\noindent
\textbf{Model settings}. We evaluate our model over four scales $n=\{1,2,3,4\}$. We use the concise visual notation:  \tempri ~, \temprii ~, \tempriii ~, \tempriv~  to refer to these 4 configurations.
Except during ablations, we follow model configurations similar to \cite{jaegle2021perceiver,lee2019set} for each attention tower ($L=8, \; d=256, \; H_{C}=4, \; H_{S}=8$)\footnote{$L$: number of self-attention layers, $d$: size of the latent bottleneck, $H_{C}$ and $H_{S}$: numbers of cross and self-attention heads respectively.}. 
We sample $F=16$ frames for each scale\footnote{We use adaptive average pooling for down-scaling encoder output features $\mathbf{z}_{i}$ across datasets to a fixed size of t=16, h=4, and w=4}.

Overall, we employ four encoder architectures. MoViNet-A4 \cite{kondratyuk2021movinets} is used for UCF-101, SSsub21 and NTU-RGB in \Cref{sec:experiments:sub::results} due to its efficiency and high accuracy on action recognition. A 3D ResNet-18 with TemPr \tempriv \ is used to compare against models with the same feature encoder in \Cref{sec:experiments:sub::results} and for the ablation studies in \Cref{sec:experiments:sub::ablation}. We additionally experiment with the widely used encoder networks, SlowFast-R50 \cite{feichtenhofer2019slowfast} for EK-100 and (video) Swin-B \cite{liu2021video} on SSv2. All convolutional encoders are pre-trained on Kinetics-700~\cite{smaira2020short} and then trained on each dataset over the full videos. Swin-B is initialized with the official weights pre-trained on Kinetics-600~\cite{carreira2018short}.

% training specifications
\noindent
\textbf{Training scheme}. For UCF-101, EK-100, and NTU-RGB, we process the videos by scaling the height to 384px and taking a center crop to size $384 \! \times \! 384$px followed by a random crop of $224 \! \times \! 224$px. Because of SSsub21's low frame resolution, we scale the input frames to $100 \! \times \! 176$px. We initialize $\beta$ with 0.5 and train for 60 epochs with $1e^{-2}$ base learning rate for TemPr and $1e^{-3}$ for $\beta$. Both learning rates are reduced on epochs $\{14, 32, 44\}$ by $1e^{-1}$. We use batch sizes of 32 for UCF-101, EK-100, NTU-RGB \& SSv2 and 64 for SSsub21 with AdamW \& $1e^{-5}$ weight decay.

\subsection{Comparative Results}
\label{sec:experiments:sub::results}

\begin{table*}[ht]
\caption{\textbf{Top-1 accuracy (\%) of EAP} over different observation ratios ($\mathbf{\rho}$). }
\begin{subtable}[c]{0.32\textwidth}
\subcaption{\textbf{NTU-RGB}.}
\label{tab:nturgb_sota}
\centering
\resizebox{\textwidth}{!}{%
\begin{tabular}{l| c c c c c c}
\hline
\multicolumn{1}{c|}{\multirow{2}{*}{Method}} &
\multicolumn{6}{c}{Observation ratios ($\rho$)} \tstrut \\
&
0.1 &
0.2 &
0.3 &
0.5 &
0.7 &
0.9 \bstrut \\
\hline
RankLSTM \cite{ma2016learning} &
11.5 & 
16.5 & 
25.7 & 
48.0 & 
61.0 & 
66.1 \tstrut \\[0.1em]

DeepSCN \cite{kong2017deep} &
16.8 & 
21.5 & 
30.6 &
48.8 & 
58.2 & 
60.0 \\[0.1em]

MSRNN \cite{hu2018early} &
15.2 &
20.3 &
29.5 &
51.6 &
63.9 &
68.9 \\[0.1em]

TS (2$\! \times \!$L) \cite{wang2019progressive} &
27.8 &
35.8 &
46.3 &
67.4 &
77.6 &
81.5 \bstrut \\
\hline
\textbf{TemPr} \tempriv \textbf{(ours)} &
\textbf{29.3} &
\textbf{38.7} &
\textbf{50.2} &
\textbf{70.1} &
\textbf{78.8} &
\textbf{84.2} \tstrut \\

\end{tabular}
}
\end{subtable}
\hfill
\begin{subtable}[c]{0.32\textwidth}
\vspace*{4pt}
\subcaption{\textbf{SSsub21}.} %with MoViNet-A4 encoder
\label{tab:sthng_smthng_sota_coarse}
\centering
\resizebox{\textwidth}{!}{%
\begin{tabular}{ l| c c c c c c }
\hline
\multicolumn{1}{c|}{\multirow{2}{*}{Method}} &
\multicolumn{6}{c}{Observation ratios ($\rho$)} \tstrut \\[.3em]
&
0.1 &
0.2&
0.3 &
0.5 &
0.7 &
0.9 \bstrut \\
\hline
mem-LSTM\cite{kong2018action} &
14.9 &
17.2 &
18.1 &
20.4 &
23.2 &
24.5  \tstrut \\[.1em]
MS-LSTM \cite{sadegh2017encouraging} &
16.9 &
16.6 &
16.8 &
16.7 &
16.9 &
17.1 \\[.1em]
MSRNN \cite{sadegh2017encouraging} &
20.1 &
20.5 &
21.1 &
22.5 &
24.0 &
27.1 \\[.1em]
GGN \cite{wu2021anticipating} &
21.2 &
21.5 &
23.3 &
27.4 &
30.2 &
30.5 \\[.1em]
IGGN \cite{wu2021spatial} &
22.6 &
- &
25.0 &
28.3 &
32.2 &
34.1 \bstrut \\
\hline
\textbf{TemPr} \tempriv \textbf{(ours)} &
\textbf{28.4} &
\textbf{34.8} &
\textbf{37.9} &
\textbf{41.3} &
\textbf{45.8} &
\textbf{48.6} \tstrut \\
\end{tabular}
}
\end{subtable}
\hfill
\begin{subtable}[c]{0.32\textwidth}
\vspace{-1.2em}
\subcaption{\textbf{SSv2}.}%  with Swin-B encoder
\label{tab:sthng_smthng_v2}
\centering
\resizebox{\textwidth}{!}{%
\begin{tabular}{l| l l l l}
\hline
\multicolumn{1}{c|}{\multirow{2}{*}{Method}} &
\multicolumn{4}{c}{Obs. ratios ($\rho$)} \tstrut \\
&
0.1 &
0.3 &
0.5 &
0.7 \bstrut \\
\hline
Baseline (Inference)  &
6.9 &
17.6 &
28.9 &
36.0 \tstrut \\[.3em]
Baseline (Fine-tuned)  &
14.4 &
23.5 &
31.1 &
39.6 \bstrut \\
\hline
\textbf{TemPr} \tempriv \  \textbf{(ours)} &
\textbf{20.5} &
\textbf{28.6} &
\textbf{41.2} &
\textbf{47.1} \tstrut \\
\end{tabular}
}
\end{subtable}
\newline
\begin{subtable}[c]{\textwidth}
\vspace*{.8em}
\subcaption{\textbf{EK-100}.} %with SlowFast-R50 encoder 
\label{tab:epic_sota_nouns}
\vspace{.2em}
\centering
\resizebox{\textwidth}{!}{%
\begin{tabular}{l| c c c c c c| c c c c c c | c c c c c c }
\hline
\multicolumn{1}{c|}{\multirow{3}{*}{Method}} &
\multicolumn{6}{c|}{Verb} &
\multicolumn{6}{c|}{Noun} &
\multicolumn{6}{c}{Action} \tstrut \bstrut \\\cline{2-19}
&
\multicolumn{18}{c}{Observation ratios ($\rho$)} \tstrut \\
&
0.1 &
0.2 &
0.3 &
0.5 &
0.7 &
0.9 &
0.1 &
0.2 &
0.3 &
0.5 &
0.7 &
0.9 &
0.1 &
0.2 &
0.3 &
0.5 &
0.7 &
0.9 \bstrut \\
\hline
Baseline (Inference) &
17.3 & 
19.7 & 
27.0 & 
48.7 & 
60.5 &
64.2 &  
19.5 &
21.7 & 
25.3 & 
38.5 &  
46.7 & 
49.1 &
5.4 &
7.6 & 
11.1 &
24.3 &
34.1 &
37.6 \tstrut \\
Baseline (Fine-tuned) &
20.6 &  
21.8 & 
29.4 & 
49.8 & 
61.3 &
64.3 &  
21.3 & 
24.2 & 
27.6 & 
39.4 &
47.3 & 
49.1 &
6.9 & 
9.1 & 
12.8 &
25.5 &
34.9 &
37.5 \bstrut \\
\hline
\textbf{TemPr} \tempriv \textbf{(ours)} &
\textbf{21.4} &  
\textbf{22.5} & 
\textbf{34.6} & 
\textbf{54.2} & 
\textbf{63.8} &
\textbf{67.0} &  
\textbf{22.8} & 
\textbf{25.5} & 
\textbf{32.3} & 
\textbf{43.4} &
\textbf{49.2} &  
\textbf{53.5} & 
\textbf{7.4} & 
\textbf{9.8} & 
\textbf{15.4} &
\textbf{28.9} &
\textbf{37.3} &
\textbf{40.8} \tstrut \\
\end{tabular}
}
\end{subtable}
\end{table*}

% UCF-101
\noindent
\textbf{UCF-101}. For a fair comparison to prior methods, we structure our results based on the feature encoder.
In the top half of \Cref{tab:ucf101_sota}, we demonstrate that our TemPr \tempriv \ model consistently outperforms all other methods with the same ResNet-18 encoder~\cite{cao2013recognize,hu2018early,kong2018action, kong2014discriminative,kong2017deep,wu2021anticipating}, for every observation ratio. Across our tests, the largest improvements are observed in small ratios in which, we achieve +8.4\% improvement for $\rho \! = \! 0.1$, +3.0\% for $\rho \! = \! 0.2$ and +2.2\% for $\rho \! = \! 0.3$, compared to the previous top-performing models. 

We also outperform prior works~\cite{ gammulle2019predicting,wu2021anticipating,fernando2021anticipating,pang2019dbdnet,wu2021spatial} on the same backbone for every $\rho$. Our method does not outperform~\cite{foo2022era} on the 48M parameters ResNeXt101 backbone. 
However, using the more efficient MoViNet-A4 or X3D$_M$ networks with 5M and 4M parameters respectively, we outperform~\cite{foo2022era} in all but $\rho=0.1$.
We get best performance of TemPr~\tempriv\ when using MoViNet-A4; e.g. at $\rho=0.3$ we outperform all prior work by $2.5\%$. 
For $\rho=0.1$, we speculate that methods like~\cite{foo2022era} benefit from specializing to subtle differences when only a handful of frames are observed. The final three rows of \Cref{tab:ucf101_sota} present results across for $n=1$, $2$ and $3$. 
Results steadily increase, across observation ratios as more scales are incorporated in TemPr.
Further results are available in \textsection \textcolor{red}{S1} in Supplementary Material.

% NTU-RGB
\noindent
\textbf{NTU-RGB}. Results on NTU-RGB are presented in \Cref{tab:nturgb_sota}. Compared to the state-of-the-art, our TemPr \tempriv \ consistently outperforms other models across the six observation ratios used. We observe the largest improvement in accuracy over \cite{wang2019progressive} at $\rho=0.3$ with 3.9\%. For smaller observation ratios, accuracy increases by 1.5\%  and 2.9\% for $\rho=0.1$ and $\rho=0.2$, respectively.

% sub-21
\noindent
\textbf{Something-Something (sub21)}. \Cref{tab:sthng_smthng_sota_coarse} demonstrates the SSsub21 class-averaged accuracy, across observation ratios as in \cite{wu2021spatial,wu2021anticipating}. Our proposed TemPr \tempriv \ surpasses state-of-the-art models~\cite{wu2021spatial,wu2021anticipating} with a significant improvement over all observation ratios. Compared to the previous top-performing model per observation ratio, accuracy increases include 5.8\% at $\rho = 0.1$, 13.3\% at $\rho=0.2$, 13.6\% at $\rho = 0.7$, and 14.5\% at $\rho=0.9$. 

% SSV2
\noindent
\textbf{Something-Something (SSv2)}.
\Cref{tab:sthng_smthng_v2} shows results on SSv2 per observation ratio with video Swin-B, which achieves 66.3\%  when evaluated on full videos (i.e. $\rho = 1.0$)\footnote{We note that the difference from the reported 69.6\% accuracy in \cite{liu2021video} is due to our use of 16 frames instead of the reported 32 frames as input.}.
We note the significant drop in performance when evaluated on partially-observed videos. Even when $\rho=0.7$, the model can only achieve 36.0\% top-1 accuracy. The improvement remains modest when the classifier is fine-tuned.
On average, TemPr \tempriv \ outperforms the inference-only model by $12.0\%$ and the fine-tuned model by $7.2\%$. Improvements are also evident across $\rho$. This not only demonstrates the benefits of our proposed TemPr model for EAP, but also the distinction between the tasks of action classification and EAP, and thus the need for EAP-specific models.

\begin{table*}[t]
\caption{\textbf{Ablation studies on UCF-101} with TemPr \tempriv \ across obs. ratios. We use $\spadesuit$ to denote softmax during training and $\clubsuit$ for $\theta = \frac{1}{2 n}$.}
\begin{subtable}[t]{0.32\textwidth}
\subcaption{\textbf{Video Scales Strategy}. }
\label{tab:ucf101_ablate_segment}
\centering
\vspace{7.5pt}
\resizebox{\textwidth}{!}{%
\begin{tabular}{l| c c c c }
\hline
\multirow{2}{*}{Scale strategy} &
\multicolumn{4}{c}{Observation ratios ($\rho$)} \tstrut \\[.2em]
&
0.2 &
0.4 &
0.6 &
0.8 \bstrut \\
\hline
full \full \  &
86.4 &
88.3 &
88.8 &
89.0 \tstrut \\[.1em]
equal \equal \  & 
83.7 &
84.6 &
86.3 &
87.1 \\[.1em]
random \random \  & 
88.8 &
89.7 &
90.2 &
90.6 \bstrut \\
\hline
decreasing \decreasing \  & 
90.0 &
\textbf{90.9} &
91.6 &
\textbf{92.6} \tstrut \\[.1em]
\textbf{increasing} \increasing \  &
\textbf{90.2} &
\textbf{90.9} &
\textbf{91.8} &
92.3  \\
\end{tabular}
}
\end{subtable}
\hfill
\begin{subtable}[t]{0.23\textwidth}
\subcaption{\textbf{Aggregation function}.}
\label{tab:ucf101_ablate_ensemble}
\centering
\vspace{7.5pt}
\resizebox{\textwidth}{!}{%
\begin{tabular}{l| c c }
\hline
\multicolumn{1}{c|}{\multirow{2}{*}{Aggregation}} &
\multicolumn{2}{c}{$\rho$} \tstrut \\[.3em]
&
0.2 &
0.4 \bstrut \\
\hline
avg & 
89.5 &
90.1 \tstrut \\[.2em]
softmax &
87.8 &
89.4 \\[.2em]
top$^\spadesuit$ &
84.6 &
87.5 \\[.2em]
gate ($\theta \! = \! 0.1$) &
85.4 &
88.5 \\[.2em]
ICW &
89.7 &
90.1 \\[.2em]
weighted &
88.5 &
89.0 \\[.2em]
weighted ($\theta$) $^\clubsuit$ & 
83.4 &
85.8 \bstrut \\
\hline
\textbf{adaptive} ($\mathcal{E}(\cdot)$) &
\textbf{90.2} &
\textbf{90.9} \tstrut \\
\end{tabular}
}
\end{subtable}
\hfill
\begin{subtable}[t]{0.24\textwidth}
\vspace{-2pt}
\subcaption{\textbf{Weight sharing} over attention towers and classifiers.}
\begin{threeparttable}[H]
\label{tab:ucf101_ablate_wsharing}
\centering
\resizebox{\textwidth}{!}{%
\begin{tabular}{c c| c c c }
\hline
\multicolumn{2}{c |}{Weight sharing} &
\multicolumn{3}{c}{$\rho$} \tstrut \\[.3em]
MAB &
$f(\cdot)$ &
0.2 &
0.4 &
0.6 
\bstrut \\
\hline
\cmark &
\xmark & 
73.4 &
76.2 &
79.0 
\tstrut \\[.1em]
\xmark &
\xmark &
84.7 &
85.8 &
87.3  
\bstrut \\
\hline
\cmark &
\cmark & 
89.2 &
90.0 &
90.7 
\tstrut \\[.1em]
\xmark &
\cmark &
\textbf{90.2} &
\textbf{90.9} &
\textbf{91.8}\\  
\end{tabular}
}
\end{threeparttable}%
\vspace{-5em}
\end{subtable}
\hfill
\begin{subtable}[r]{.18\textwidth}
\vspace{-4pt}
\subcaption{\textbf{Latent array ($\mathbf{u}$) sharing}.}
\label{tab:ucf101_ablate_share_latent}
\resizebox{\textwidth}{!}{%
\begin{tabular}{c| c | c c }
\hline
\multicolumn{1}{c|}{\multirow{2}{*}{$\mathbf{u}$}} &
Mem. &
\multicolumn{2}{c}{$\rho$} \tstrut \\[.3em]
shared &
(GB) &
0.2 &
0.4 \bstrut \\
\hline
\xmark & 
4.0 &
\textbf{90.2} &
\textbf{91.0} \tstrut \bstrut \\[.2em]
\hline
\cmark &
\textbf{3.0} &
\textbf{90.2} &
90.9 \tstrut \\[.2em]
\end{tabular}
}
\vspace{-70pt}
\end{subtable}
\newline
\begin{subtable}[r]{.23\textwidth}
\hspace{15pt}
\subcaption{\textbf{CMAB replacements.}}
\label{tab:ucf101_attention}
\centering
\vspace{1pt}
\resizebox{\linewidth}{!}{%
\begin{tabular}{l| l l | c c}
\hline
\multicolumn{1}{c|}{\multirow{2}{*}{MAB}} &
\multicolumn{2}{c|}{$\rho$} &
\parbox[t]{1mm}{\multirow{3}{*}{\rotatebox[origin=c]{90}{\small{ Par. (M)}}}} &
\parbox[t]{1mm}{\multirow{3}{*}{\rotatebox[origin=c]{90}{\small{GFLOPs }}}}
\rule{0pt}{3.4ex} \\[.8em]
&
0.2 &
0.4 & 
&
\bstrut \\[.4em]
\hline
Self &
83.2 &
84.5 &
84.6 &
8.59  \tstrut \bstrut \\
\hline
\textbf{Cross} &
\textbf{90.2} &
\textbf{90.9} &
23.0 & 
1.47  \tstrut \\
\end{tabular}
}
\vspace{1.5em}
\end{subtable}
\hspace{5em}
\vspace*{1.4cm}
\vspace{-4em}
\end{table*}

% EPIC KITCHENS
\noindent
\textbf{EPIC-KITCHENS-100 (EK-100)}. We also investigate EAP on EK-100. We believe that a challenging part of EK-100 is the inclusion of fine-grained verb labels. For example, the class `hold' is easily confused with partially-observed videos of classes `put', `throw', `insert' or `stack'. These classes start with objects being held before the action is initiated. We are the first to use EK-100 as a benchmark for EAP. As in SSv2, we report inference-only and classifier fine-tuned models alongside TemPr \tempriv -. 

\Cref{tab:epic_sota_nouns} demonstrates the performance per observation ratio. 
TemPr~\tempriv\ outperforms the baselines and showcases that EK-100 is more challenging than all other benchmarks when focusing on action performance - $28.9\%$ for $\rho=0.5$ compared to $95.4\%, 70.1\%$ and $41.2\%$ for UCF-101, NTU-RGB, and SSv2.
We note that EAP is higher for noun classes in smaller $\rho$ while classifying verbs becomes easier for larger $\rho$. This highlights that actions, which require correct prediction of the verb and the noun, are challenging to be predicted in cases where very few frames are observed.

\subsection{Ablation Studies and Qualitative Results}
\label{sec:experiments:sub::ablation}

% Overview
In this section we conduct ablation studies on UCF-101 reporting accuracy over different observation ratios.
Unless specified, we use the ResNet-18 backbone.
Computations and memory use are reported solely for TemPr, without the encoder, to demonstrate the differences clearer.

% Scales analysis
\noindent
\textbf{Video scales strategy}. Different strategies can be used for selecting video scales. We compare our proposed temporal progressive sampling (\Cref{sec:method:sub::sampling}) to other common strategies and potential baselines in \Cref{tab:ucf101_ablate_segment}.
In all settings, we keep $n=4$ scales.
The \textit{full} strategy \full~ uses $n$ scales of fixed length matching the entire observation video. In \textit{equal} \equal{}, scales/segments have equal lengths as in \cite{wang2016temporal}. The \textit{random} strategy \random~uses scales of random length. Finally, the \textit{increasing} \increasing{} and \textit{decreasing} \decreasing{} strategies utilize our proposed progressive approach, sampling the fine scale from either the start or the end of the observed video. Accuracy is consistently lower when scales are of the same length, either matching the observed video (\textit{full}) or equally-sized (\textit{equal}). This is in contrast to the success of this sampling approach for action recognition~\cite{wang2016temporal}, further emphasizing the distinction between the two tasks. The use of progressive (\textit{increasing} or \textit{decreasing}) video scales exhibits an average +3.6\% accuracy increase across $\rho$, compared to other sampling approaches. We note that no model component depends on the order of the scales, thus the performance over increasing or decreasing scales is expected to be similar. 

In \Cref{tab:sssub21_scale_strategy}, we compare sampling strategies on SSsub21, as this dataset is more challenging temporally. We use TemPr \tempriv \ with MoViNet-A4. Similar to \Cref{tab:ucf101_ablate_segment}, progressive (\textit{increasing} \increasing~ or \textit{decreasing} \decreasing~) scales is a better-suited strategy, with an average +2.4\% accuracy increase over $\rho$. This emphasizes the need for fine-to-coarse sampling, independent of where the fine sample is taken from.

\begin{table}[t]
\RawFloats
\vspace{-14em}
\captionsetup{width=0.65\textwidth,justification=raggedright,singlelinecheck=false,margin=2mm}
\caption{\textbf{Video Scales Strategies on \\ SSsub21} with TemPr \tempriv \ .}
\label{tab:sssub21_scale_strategy}
\vspace{-.3em}
\resizebox{.65\textwidth}{!}{%
\begin{tabular}{l| c c c c }
\hline
\multicolumn{1}{c|}{\multirow{2}{*}{\shortstack{Scale \\ strategy}}} &
\multicolumn{4}{c}{Obs. ratios ($\rho$)} \\[.2em]
&
0.2 &
0.3 &
0.5 &
0.7 \bstrut \\[.2em]
\hline
full \full \ \tstrut &
32.6 &
36.4 &
39.3 &
42.9 \\[.1em]
equal \equal \  & 
29.8 &
34.5 &
37.2 &
41.8 \\[.1em]
random \random \  & 
33.4 &
37.1 &
40.6 &
44.3 \bstrut \\[.1em]
\hline
decreasing \decreasing \ \tstrut &
\textbf{35.2} &
\textbf{38.3} &
40.7 &
45.2 \\[.1em]
\textbf{increasing} \increasing \ &
34.8 &
37.9 &
\textbf{41.3} &
\textbf{45.8} \\[.1em]
\end{tabular}
}\hfill
\end{table}

% Ensemble methods
\noindent
\textbf{Prediction aggregation}. \Cref{tab:ucf101_ablate_ensemble} presents comparisons over different aggregation functions. In the case that the predictor with the highest confidence is chosen (top$^{\spadesuit}$), we use softmax during training to ensure that gradients are propagated across the entire network. The largest drop in performance is observed when using individual predictions (softmax, top, gate). Methods that are instead based on using all predictors uniformly by averaging them, or by weighting them with Inverse Covariance Weighting (ICW) improve the final predictions. A further +0.7\% accuracy over ICW is observed by our adaptive approach with the combination of predictor agreement and confidences.

\begin{figure}[t]
\RawFloats
\vspace{-16em}
    \begin{center}
        $\qquad \quad$
        \includegraphics[width=.52\linewidth,trim={2cm 0 2cm 0},clip]{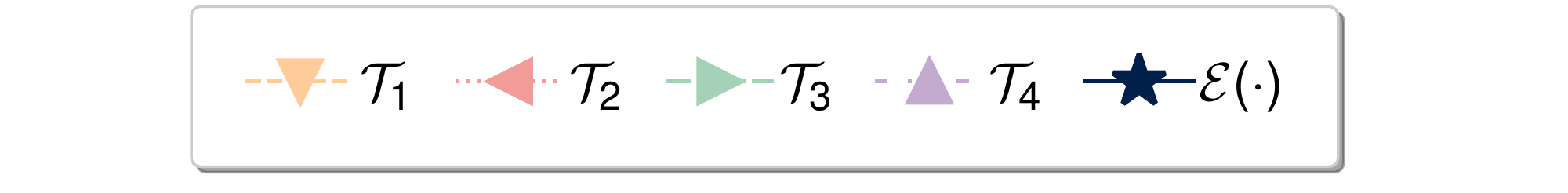}
        \vspace{-1em}
    \end{center}
  \begin{minipage}{\linewidth}
    \centering
    $\qquad$
    \subcaptionbox{\textbf{UCF-101}.}
    {\includegraphics[width=.42\linewidth,trim={0 5cm 0 0},clip]{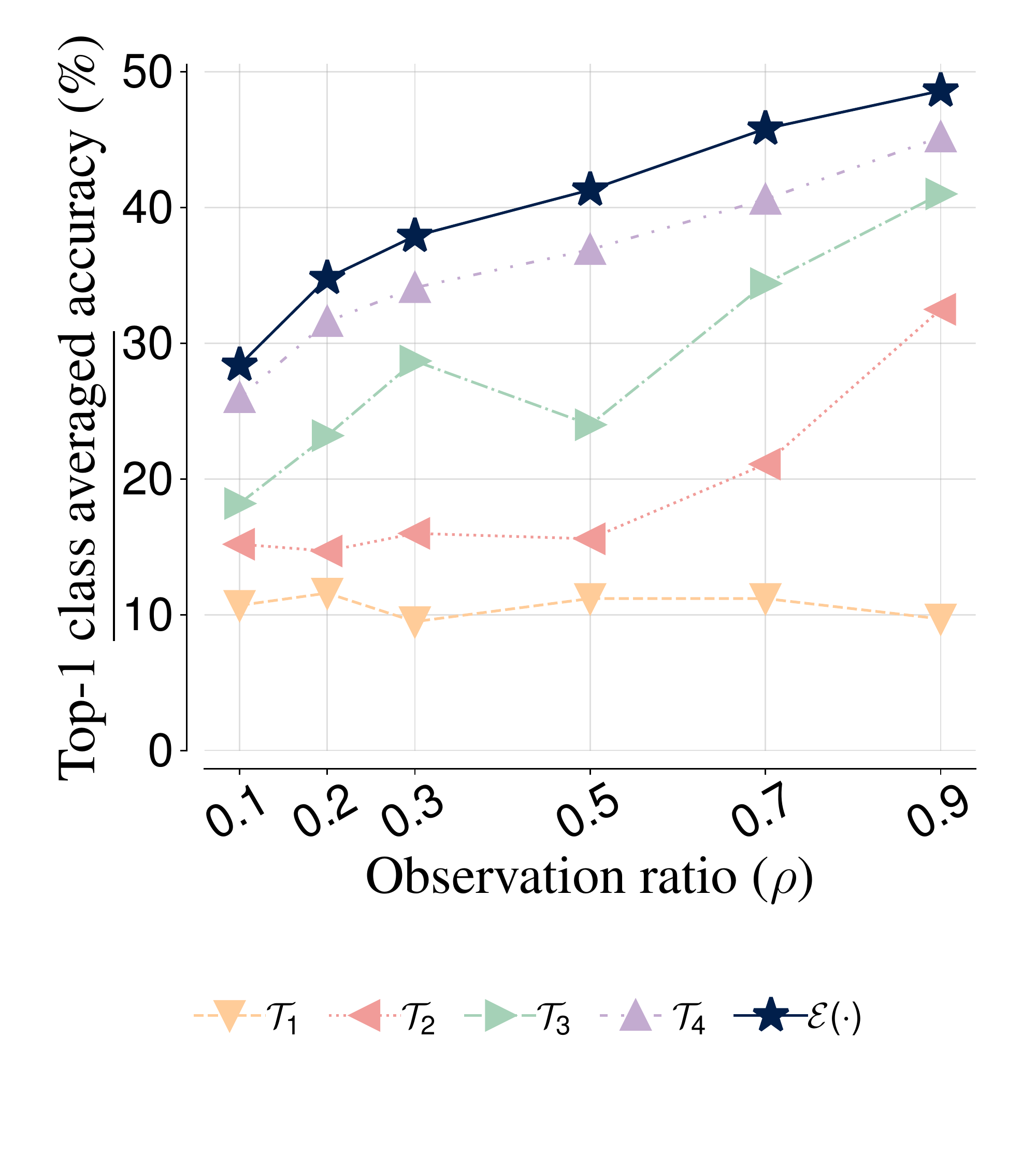}}$\;$
    \subcaptionbox{\textbf{SSsub21}.}
     {\includegraphics[width=.42\linewidth,trim={0 5cm 0 0},clip]{figures/tempr_sampler2obsratio_ss21_class_legend.pdf}\label{fig:sssub21_tower_acc}}
    \caption[Top-1 accuracy of each TemPr]{\textbf{Top-1 accuracy of each TemPr} \tempriv \ \textbf{tower} $\mathcal{T}_{i}$ \textbf{per} $\rho$.}
    \label{fig:tower_accs}
  \end{minipage}
\end{figure}

% Weight sharing results analysis
\noindent
\textbf{Weight sharing combinations}. We consider the two model components that can share their weights across scales. The first is the multihead-attention blocks~(MAB) and the second is their classifier layer. \Cref{tab:ucf101_ablate_wsharing} shows that using individual classifier weights for each tower decreases performance. 
Classifier weight sharing improves performance.

\begin{figure*}[t]
\includegraphics[width=\linewidth]{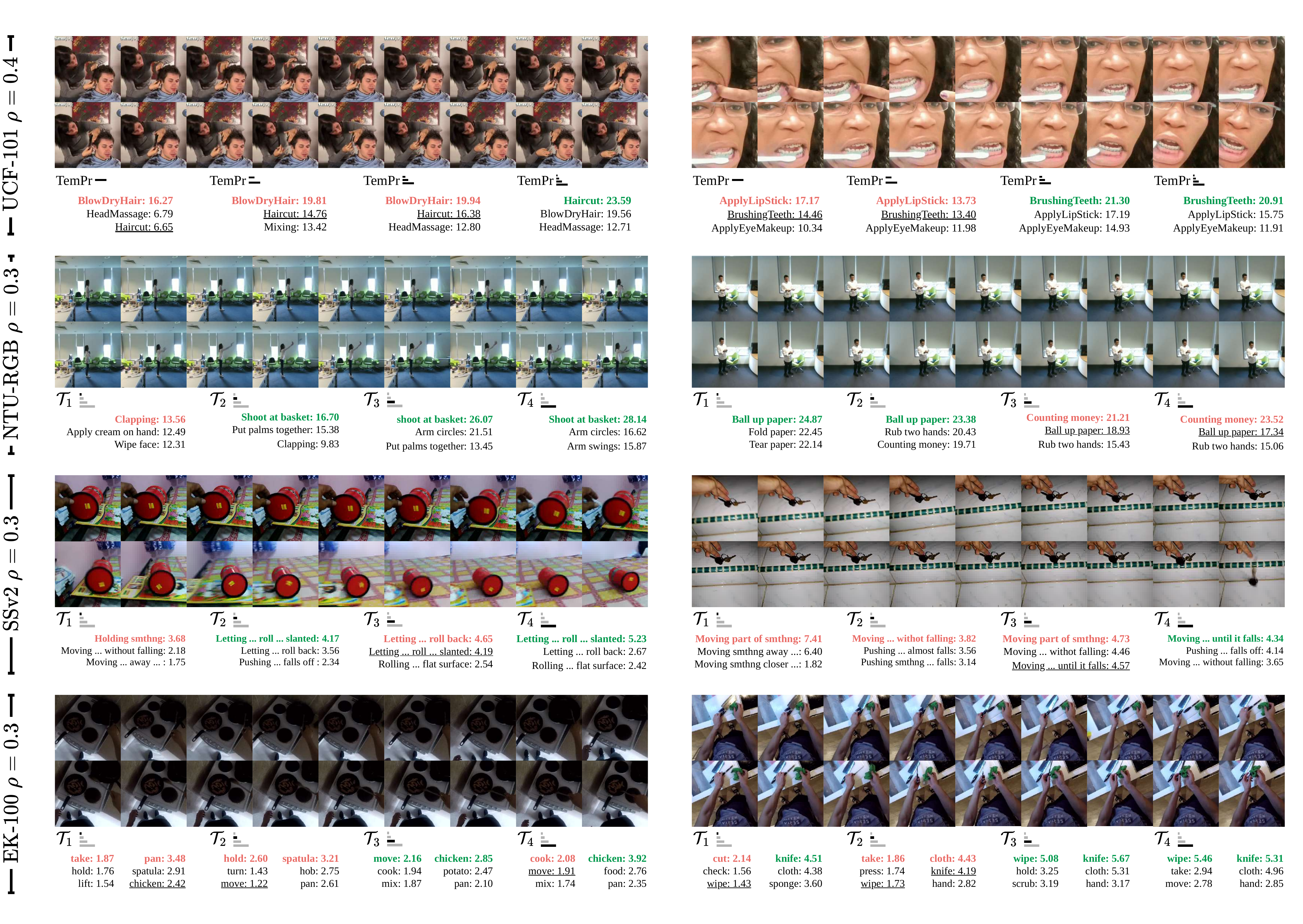}
\vspace*{4pt}
\caption{\textbf{Examples from UCF-101, NTU-RGB, SSv2 and EK-100}. Top 3 action label confidences are reported for either TemPr model or over individual tower predictors ($\mathcal{T}_i$). We show the 16 frames sampled per video. \textcolor{applegreen}{Green}/\textcolor{cadmiumred}{red} highlight correct/incorrect top 1 predictions, and we \underline{underline} true label when in top-3. We show verb and noun predictions for EK-100. See additional examples in \textsection \textcolor{red}{S6}.} 
\label{fig:tempr_acc_instances}
\vspace*{-25pt}
\end{figure*}

% Latent array weight sharing
\noindent
\textbf{Latent array ($\mathbf{u}$)}. \Cref{tab:ucf101_ablate_share_latent} shows the effect on both performance and memory when sharing the Cross MAB latent array $\mathbf{u}$ across attention towers. With marginal difference in accuracy, sharing $\mathbf{u}$ increases efficiency with a significant reduction in memory. Thus, we share $\mathbf{u}$ in all experiments.

\noindent
\textbf{CMAB replacements}. We include ablations on the effect of cross/self-MAB in accuracy, compute and memory on \Cref{tab:ucf101_attention}. We note that self-MAB-only towers significantly increase memory and computation costs.

% Scale to \rho
\noindent
\textbf{Scale per Observation Ratio}. We additionally plot the performance of individual predictors for both UCF-101 and SSsub21 in \Cref{fig:tower_accs} with respect to different observation ratios. As shown, datasets such as Something-Something that are less appearance-based can benefit more from the proposed aggregated progressive scales. Class accuracies across scales are presented in \textsection \textcolor{red}{S1}. Overall, towers of smaller scales ($\mathcal{T}_1$~\temprivbcd{} and $\mathcal{T}_2$~\temprivacd{}) performed more favorably for classes that are distinguishable from the only first few frames. In contrast, towers of larger scales ($\mathcal{T}_3$~\temprivabd{} and $\mathcal{T}_4$~\temprivabc{}) were better suited for classes that the action become distinguishable with a larger part of the video observed.

\noindent
\textbf{Qualitative results}. The first row of \Cref{fig:tempr_acc_instances} demonstrates UCF-101 instances where predictions differ across TemPr \tempri~ , \temprii~ , \tempriii~ , \tempriv~ . The increase in the number of scales allows the network to capture features that are more descriptive of the target action e.g. the two \textit{BrushingTeeth} instances. 
In the first example, the subtle motion of \textit{Hair Cutting} is only confidently predicted when the finest scale is incorporated in TemPr (comparing \tempriii~ to \tempriv~ ).
In the following three rows of \Cref{fig:tempr_acc_instances}, predictions from individual towers $\mathcal{T}_1$ \temprivbcd~ , $\mathcal{T}_2$ \temprivacd~ , $\mathcal{T}_3$ \temprivabd \ and $\mathcal{T}_4$ \temprivabc~  are shown across NTRU-RGB, SSv2, and EK-100.
In the second row, fine scales benefit subtle motion e.g. in the \textit{Ball up paper}. In the third row, coarse scales assist prediction as the end of the sequence changes the prediction to the correct class, e.g. \textit{Moving something until it falls} in SSv2.
In the fourth row, coarser scales are required to distinguish \textit{taking cloth} from \textit{wiping knife} in EK-100.

\section{Conclusions}

We have proposed to utilize progressive scales from partially observed videos for early action prediction. Based on these scales, we introduce a temporal progressive (TemPr) model consisting of bottleneck-based attention towers, in order to capture the progression of an action over multiple fine-to-coarse scales. We aggregate scale predictors considering the similarity in their probability distributions as well as their confidence. 
Extensive experiments over five encoders and four video datasets demonstrate the merits of TemPr~\tempriv\~. Additionally, we are the first to investigate the unique difficulties of EAP for large-scale datasets  - evaluating EAP on SSv2 and EK-100. We hope that our approach of progressive, rather than single continual, scales can pave a new path for subsequent methods. 

\noindent \textbf{Acknowledgments.} 
We use publicly available datasets.
Research is funded by the United Nation’s End Violence Fund (iCOP 2.0) and EPSRC UMPIRE (EP/T004991/1). We utilized Bristol's HPC Blue Crystal 4 facility.

%%%%%%%%% REFERENCES
{\small
\bibliographystyle{ieee_fullname}
\bibliography{egbib}
}

\clearpage

% Pages are numbered in submission mode, and unnumbered in camera-ready
\setcounter{section}{0}
\setcounter{equation}{0}
\setcounter{figure}{0}
\setcounter{table}{0}
\SupplementaryMaterials

\twocolumn[{%
\renewcommand\twocolumn[1][]{#1}%
\begin{center}
\textbf{\Large{The Wisdom of Crowds: Temporal Progressive Attention \\ for Early Action Prediction -- Supplementary Material\\}}
\captionof{table}{\textbf{Ablation studies across scales $n=\{1,2,3,4\}$ on UCF-101 over different observation ratios ($\mathbf{\rho}$)}. Methods are grouped w.r.t. the backbone used. The best overall performance per $\mathbf{\rho}$ is in \textbf{bold} and the second best results are \underline{underlined}.} 
\label{tab:ucf101_sota}
\centering
\resizebox{.94\textwidth}{!}{%
\begin{tabular}{l| l c | c c c c c c c c c }
\hline
\multicolumn{1}{c|}{\multirow{2}{*}{Method}} &
\multicolumn{1}{c}{\multirow{2}{*}{Backbone}} &
\multicolumn{1}{c|}{\multirow{2}{*}{dim}} &
\multicolumn{9}{c}{Observation ratios ($\rho$)} \tstrut \\
&
&
&
0.1 &
0.2 &
0.3 &
0.4 &
0.5 &
0.6 &
0.7 &
0.8 &
0.9  \tstrut \bstrut \\
\hline 

\textbf{TemPr} \tempri~ \textbf{(ours)} &
\multirow{4}{*}{X3D$_{M}$}&
\multirow{4}{*}{3D} &
84.8 & 
91.8 & 
92.3 & 
92.6 & 
93.0 & 
93.4 & 
93.5 & 
93.6 & 
93.6 \tstrut \\[.1em]

\textbf{TemPr} \temprii~ \textbf{(ours)} &
&
&
85.3 & 
92.3 & 
92.8 & 
93.7 & 
93.9 & 
93.9 & 
94.2 & 
94.4 & 
94.3 \\[.1em]

\textbf{TemPr} \tempriii~ \textbf{(ours)} &
&
&
87.4 & 
93.3 & 
93.9 & 
94.4 & 
94.0 & 
94.2 & 
94.4 & 
94.9 & 
94.9 \\[.1em]

\textbf{TemPr} \tempriv~ \textbf{(ours)} &
&
&
\underline{87.9} & 
\underline{93.4} & 
\underline{94.5} & 
\underline{94.8} & 
95.1 & 
\textbf{95.2} & 
\textbf{95.6} & 
\underline{96.4} & 
\textbf{96.3} \bstrut \\
\hline
\textbf{TemPr} \tempri~ \textbf{(ours)} &
\multirow{4}{*}{MoViNet-A4}&
\multirow{4}{*}{3D} &
85.2 & 
92.1 & 
92.5 & 
92.9 & 
93.3 & 
93.7 & 
93.5 & 
93.8 & 
93.7 \\[.1em]

\textbf{TemPr} \temprii~ \textbf{(ours)} &
&
&
85.6 & 
92.9 & 
93.6 & 
94.5 & 
94.4 & 
94.2 & 
94.2 & 
94.6 & 
94.8 \\[.1em]

\textbf{TemPr} \tempriii~ \textbf{(ours)} &
&
&
87.3 & 
93.1 & 
\textbf{94.9} & 
94.6 & 
\underline{95.2} & 
\underline{94.9} & 
94.6 & 
95.1 & 
95.0 \\[.1em]

\textbf{TemPr} \tempriv~ \textbf{(ours)} &
&
&
\textbf{88.6} & 
\textbf{93.5} & 
\textbf{94.9} & 
\textbf{94.9} & 
\textbf{95.4} & 
\textbf{95.2} & 
\underline{95.3} & 
\textbf{96.6} & 
\underline{96.2} \\
\end{tabular}
}
\label{tab:ucf101_scales}
\hfill
\end{center}%
}]

\begin{figure*}[t]
\includegraphics[width=.93\linewidth]{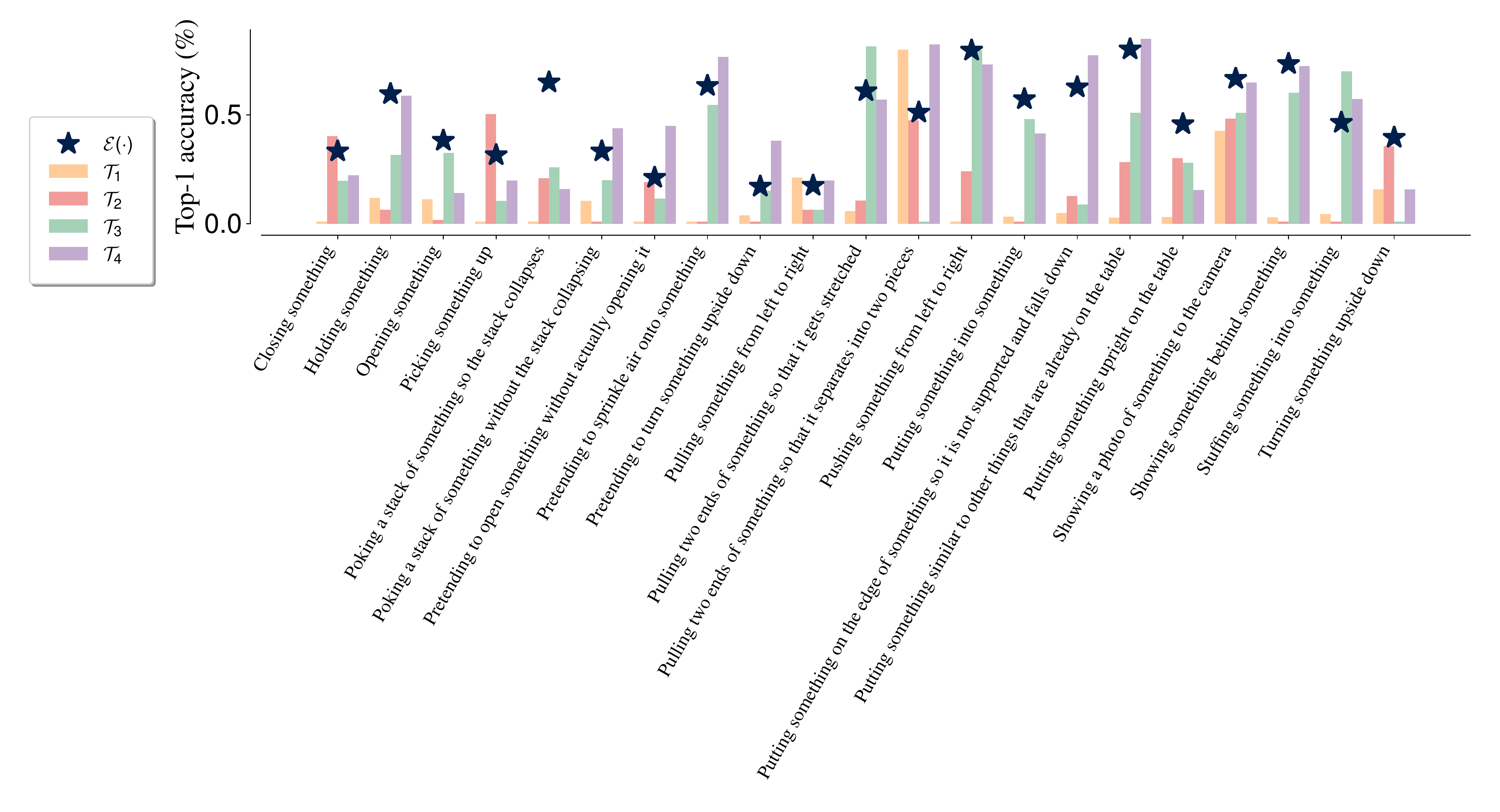}
\caption{\textbf{TemPr} \tempriv ~\textbf{SSsub21 class accuracies}  over observation ratio $\rho=0.3$.}
\label{fig:tempr_acc_class_sub21}
\end{figure*}

\section{Cross-scale accuracy and class predictions}

\begin{table}[t]
\RawFloats
\caption{\textbf{Top tower predictors per class and observation ratio for TemPr} \tempriv~. Towers \hla{$\mathcal{T}_1$} \temprivbcd~ , \hlb{$\mathcal{T}_2$} \temprivacd~ , \hlc{$\mathcal{T}_3$} \temprivbcd~   and , \hld{$\mathcal{T}_4$} \temprivabc~ are highlighted for better readability. }
\label{tab:sssub21_class_towers}
\centering
\resizebox{\linewidth}{!}{%
\begin{tabular}{|l| l l l l l l |}
\hline
\multirow{2}{*}{class name} & \multicolumn{6}{c|}{Observation ratios $\rho$}\\
 & 0.1 & 0.2 & 0.3 & 0.5 & 0.7 & 0.9 \bstrut \\
\hline
Putting smthng similar to other things ... & 
\cellcolor{a4purple} $\mathcal{T}_4$ & 
\cellcolor{a4purple} $\mathcal{T}_4$ & 
\cellcolor{a4purple} $\mathcal{T}_4$ & 
\cellcolor{a4purple} $\mathcal{T}_4$ & 
\cellcolor{a4purple} $\mathcal{T}_4$ & 
\cellcolor{a4purple} $\mathcal{T}_4$ \tstrut \\[.1em]
Showing smthng behind smthng & 
\cellcolor{a4purple} $\mathcal{T}_4$ & 
\cellcolor{a4purple} $\mathcal{T}_4$ & 
\cellcolor{a4purple} $\mathcal{T}_4$ & 
\cellcolor{a4purple} $\mathcal{T}_4$ & 
\cellcolor{a4purple} $\mathcal{T}_4$ & 
\cellcolor{a3green} $\mathcal{T}_3$ \\[.1em]
Holding smthng & 
\cellcolor{a4purple} $\mathcal{T}_4$ & 
\cellcolor{a4purple} $\mathcal{T}_4$ & 
\cellcolor{a4purple} $\mathcal{T}_4$ & 
\cellcolor{a3green} $\mathcal{T}_3$ & 
\cellcolor{a4purple} $\mathcal{T}_4$ &
\cellcolor{a4purple} $\mathcal{T}_4$ \\[.1em]
Poking ... smthng without ... collapsing & 
\cellcolor{a4purple} $\mathcal{T}_4$ & 
\cellcolor{a4purple} $\mathcal{T}_4$ &
\cellcolor{a4purple} $\mathcal{T}_4$ &
\cellcolor{a3green} $\mathcal{T}_3$ &
\cellcolor{a4purple} $\mathcal{T}_4$ &
\cellcolor{a4purple} $\mathcal{T}_4$ \\[.1em]
Pretending to sprinkle air onto smthng & 
\cellcolor{a3green} $\mathcal{T}_3$ & 
\cellcolor{a4purple} $\mathcal{T}_4$ &
\cellcolor{a4purple} $\mathcal{T}_4$ &
\cellcolor{a4purple} $\mathcal{T}_4$ &
\cellcolor{a4purple} $\mathcal{T}_4$ &
\cellcolor{a3green} $\mathcal{T}_3$\\[.1em]
Pulling two ends of smthng ... stretched & 
\cellcolor{a4purple} $\mathcal{T}_4$ & 
\cellcolor{a4purple} $\mathcal{T}_4$ & 
\cellcolor{a3green} $\mathcal{T}_3$ & 
\cellcolor{a3green} $\mathcal{T}_3$ & 
\cellcolor{a4purple} $\mathcal{T}_4$ & 
\cellcolor{a4purple} $\mathcal{T}_4$ \\[.1em]
Putting smthng into smthng & 
\cellcolor{a4purple} $\mathcal{T}_4$ & 
\cellcolor{a3green} $\mathcal{T}_3$ & 
\cellcolor{a3green} $\mathcal{T}_3$ & 
\cellcolor{a4purple} $\mathcal{T}_4$ & 
\cellcolor{a4purple} $\mathcal{T}_4$ & 
\cellcolor{a4purple} $\mathcal{T}_4$ \\[.1em]
Pretending to turn smthng upside down & 
\cellcolor{a4purple} $\mathcal{T}_4$ & 
\cellcolor{a3green} $\mathcal{T}_3$ & 
\cellcolor{a4purple} $\mathcal{T}_4$ &
\cellcolor{a3green} $\mathcal{T}_3$ &
\cellcolor{a3green} $\mathcal{T}_3$ &
\cellcolor{a4purple} $\mathcal{T}_4$ \\[.1em]
Poking a stack of smthng ... collapses & 
\cellcolor{a4purple} $\mathcal{T}_4$ & 
\cellcolor{a4purple} $\mathcal{T}_4$ & 
\cellcolor{a3green} $\mathcal{T}_3$ &
\cellcolor{a4purple} $\mathcal{T}_4$ &
\cellcolor{a3green} $\mathcal{T}_3$ &
\cellcolor{a3green} $\mathcal{T}_e$ \\[.1em]
Pulling smthng from left to right & 
\cellcolor{a3green} $\mathcal{T}_3$ &
\cellcolor{a4purple} $\mathcal{T}_4$ &
\cellcolor{a4purple} $\mathcal{T}_4$ &
\cellcolor{a3green} $\mathcal{T}_3$ & 
\cellcolor{a4purple} $\mathcal{T}_4$ &
\cellcolor{a3green} $\mathcal{T}_3$ \\[.1em]
Pushing smthng from left to right  & 
\cellcolor{a3green} $\mathcal{T}_3$ & 
\cellcolor{a4purple} $\mathcal{T}_4$ & 
\cellcolor{a3green} $\mathcal{T}_3$ & 
\cellcolor{a3green} $\mathcal{T}_3$ & 
\cellcolor{a4purple} $\mathcal{T}_4$ & 
\cellcolor{a4purple} $\mathcal{T}_4$\\[.1em]
Pretending to open smthng without ... & 
\cellcolor{a4purple} $\mathcal{T}_4$ & 
\cellcolor{a3green} $\mathcal{T}_3$ &
\cellcolor{a4purple} $\mathcal{T}_4$ &
\cellcolor{a3green} $\mathcal{T}_3$ &
\cellcolor{a3green} $\mathcal{T}_3$ &
\cellcolor{a2red} $\mathcal{T}_2$ \\[.1em]
Opening smthng & 
\cellcolor{a4purple} $\mathcal{T}_4$ &
\cellcolor{a4purple} $\mathcal{T}_4$ &
\cellcolor{a3green} $\mathcal{T}_3$ &
\cellcolor{a3green} $\mathcal{T}_3$ &
\cellcolor{a2red} $\mathcal{T}_2$ & 
\cellcolor{a2red} $\mathcal{T}_2$  \\[.1em]
Showing a photo of smthng ... & 
\cellcolor{a4purple} $\mathcal{T}_4$ & 
\cellcolor{a3green} $\mathcal{T}_3$ & 
\cellcolor{a4purple} $\mathcal{T}_4$ & 
\cellcolor{a2red} $\mathcal{T}_2$ & 
\cellcolor{a2red} $\mathcal{T}_2$ &
\cellcolor{a1orange} $\mathcal{T}_1$ \\[.1em]
Stuffing smthng into smthng & 
\cellcolor{a4purple} $\mathcal{T}_4$ & 
\cellcolor{a3green} $\mathcal{T}_3$ & 
\cellcolor{a3green} $\mathcal{T}_3$ & 
\cellcolor{a2red} $\mathcal{T}_2$ & 
\cellcolor{a2red} $\mathcal{T}_2$ & 
\cellcolor{a2red} $\mathcal{T}_2$ \\[.1em]
Putting smthng on the edge of smthng ... & 
\cellcolor{a4purple} $\mathcal{T}_4$ & 
\cellcolor{a3green} $\mathcal{T}_3$ & 
\cellcolor{a4purple} $\mathcal{T}_4$ & 
\cellcolor{a2red} $\mathcal{T}_2$ & 
\cellcolor{a1orange} $\mathcal{T}_1$ & 
\cellcolor{a1orange} $\mathcal{T}_1$ \\[.1em]
Picking smthng up & 
\cellcolor{a4purple} $\mathcal{T}_4$ & 
\cellcolor{a3green} $\mathcal{T}_3$ &
\cellcolor{a2red} $\mathcal{T}_2$ &
\cellcolor{a2red} $\mathcal{T}_2$ &
\cellcolor{a1orange} $\mathcal{T}_1$ &
\cellcolor{a2red} $\mathcal{T}_2$ \\[.1em]
Closing smthng & 
\cellcolor{a4purple} $\mathcal{T}_4$ & 
\cellcolor{a3green} $\mathcal{T}_3$ & 
\cellcolor{a2red} $\mathcal{T}_2$ & 
\cellcolor{a2red} $\mathcal{T}_2$ & 
\cellcolor{a3green} $\mathcal{T}_3$ & 
\cellcolor{a2red} $\mathcal{T}_2$ \\[.1em]
Putting smthng upright on the table & 
\cellcolor{a4purple} $\mathcal{T}_4$ & 
\cellcolor{a3green} $\mathcal{T}_3$ & 
\cellcolor{a2red} $\mathcal{T}_2$ &
\cellcolor{a1orange} $\mathcal{T}_1$ &
\cellcolor{a2red} $\mathcal{T}_2$ & 
\cellcolor{a2red} $\mathcal{T}_2$ \\[.1em]
Turning smthng upside down & 
\cellcolor{a3green} $\mathcal{T}_3$ & 
\cellcolor{a3green} $\mathcal{T}_3$ & 
\cellcolor{a2red} $\mathcal{T}_2$ & 
\cellcolor{a2red} $\mathcal{T}_2$ & 
\cellcolor{a2red} $\mathcal{T}_2$ & 
\cellcolor{a1orange} $\mathcal{T}_1$\\ [.1em]
Pulling two ends of smthng ... two pieces & 
\cellcolor{a3green} $\mathcal{T}_3$ & 
\cellcolor{a2red} $\mathcal{T}_2$ & 
\cellcolor{a1orange} $\mathcal{T}_1$ & 
\cellcolor{a2red} $\mathcal{T}_2$ & 
\cellcolor{a2red} $\mathcal{T}_2$ & 
\cellcolor{a2red} $\mathcal{T}_2$\\[.1em]
\hline
\end{tabular}
}
\end{table}

\begin{figure}[ht]
\centering
\includegraphics[width=.8\linewidth]{figures/class2tower_accuracies/legend.pdf}
\begin{subfigure}[b]{0.47\linewidth}
   \includegraphics[width=\linewidth]{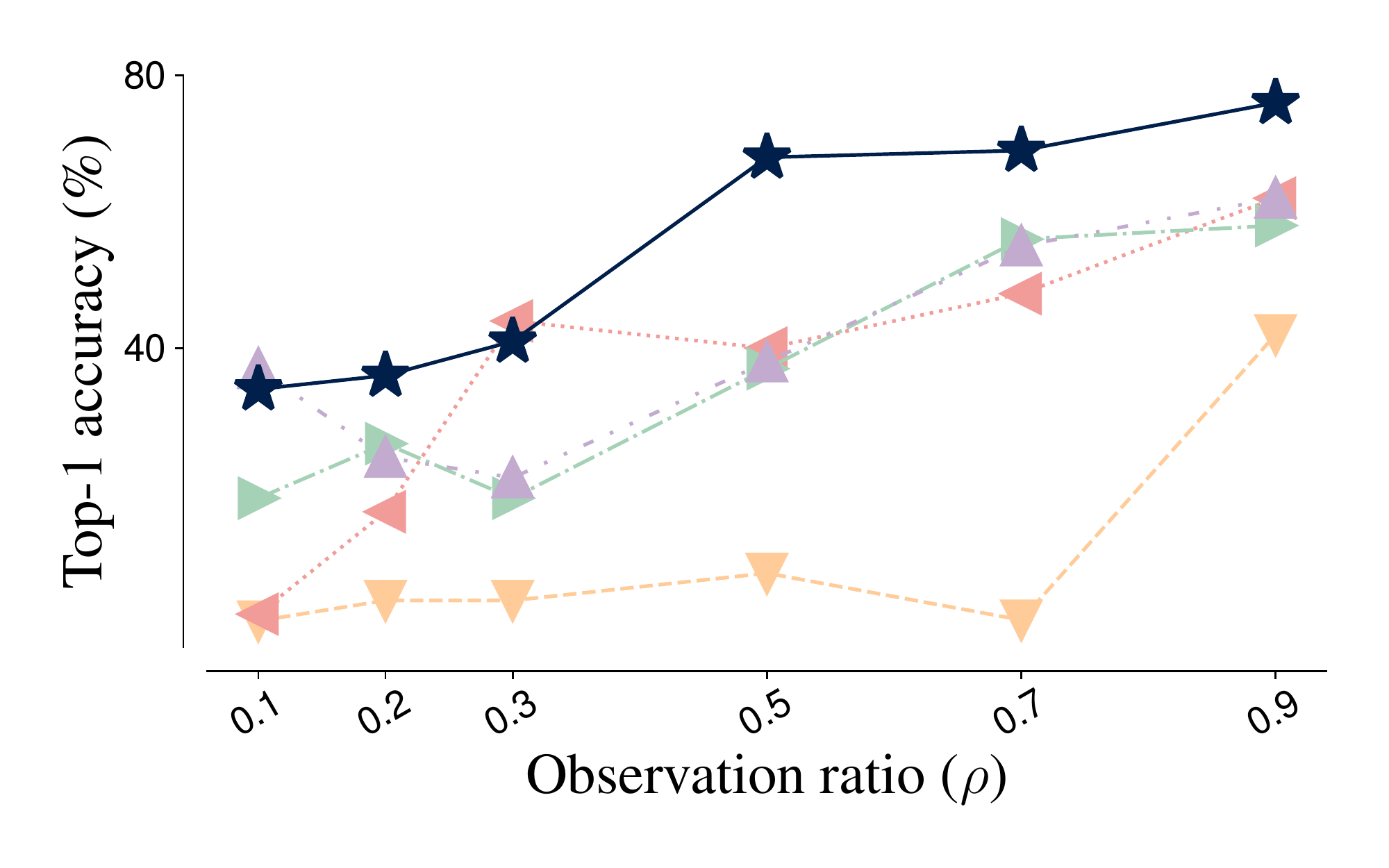}
   \caption{\textit{Closing Something}}
   \label{fig:closing_something} 
\end{subfigure}
\begin{subfigure}[b]{0.47\textwidth}
   \includegraphics[width=\linewidth]{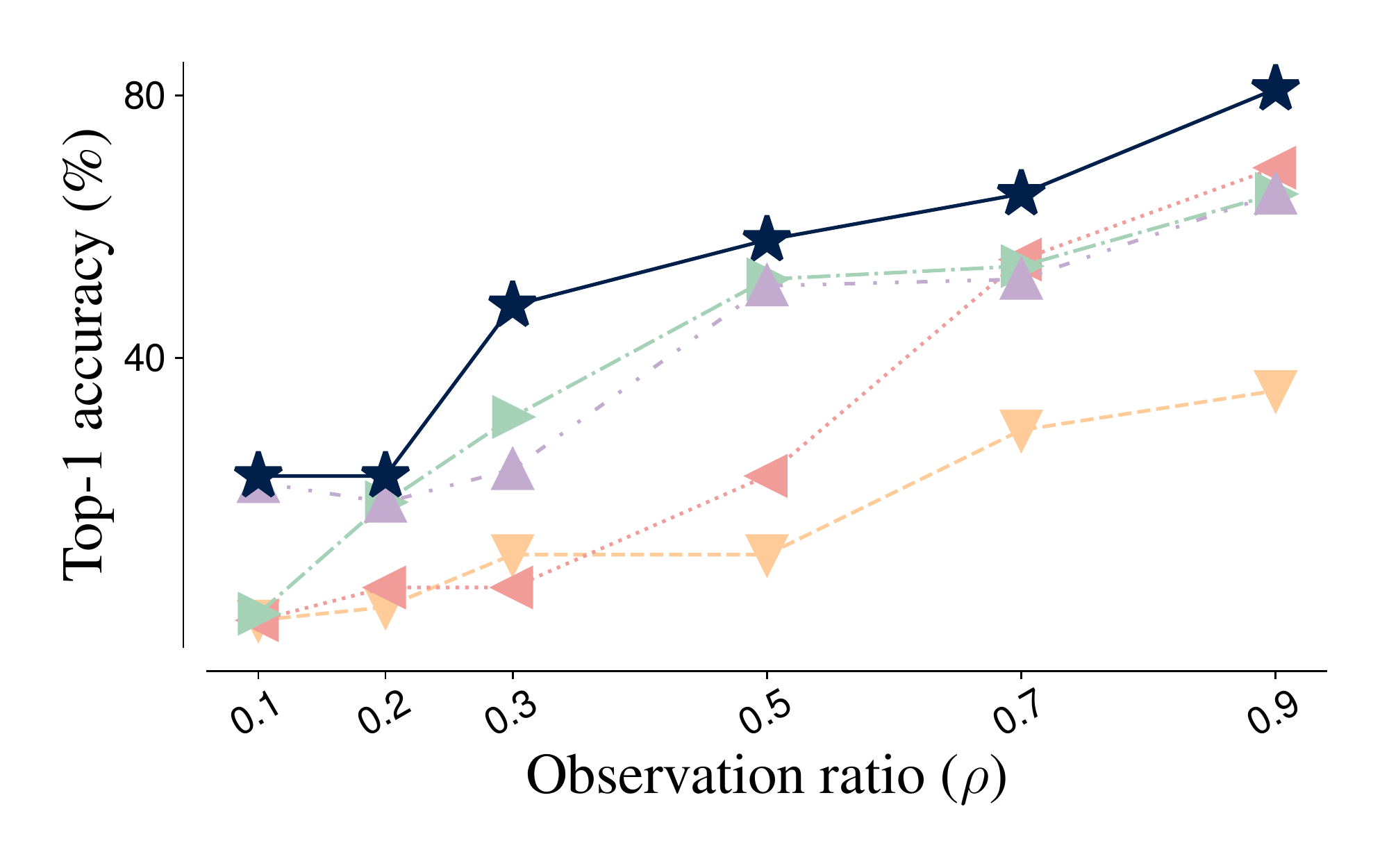}
   \caption{\textit{Opening Something}}
   \label{fig:opening_something}
\end{subfigure}
\begin{subfigure}[b]{0.47\textwidth}
   \includegraphics[width=\linewidth]{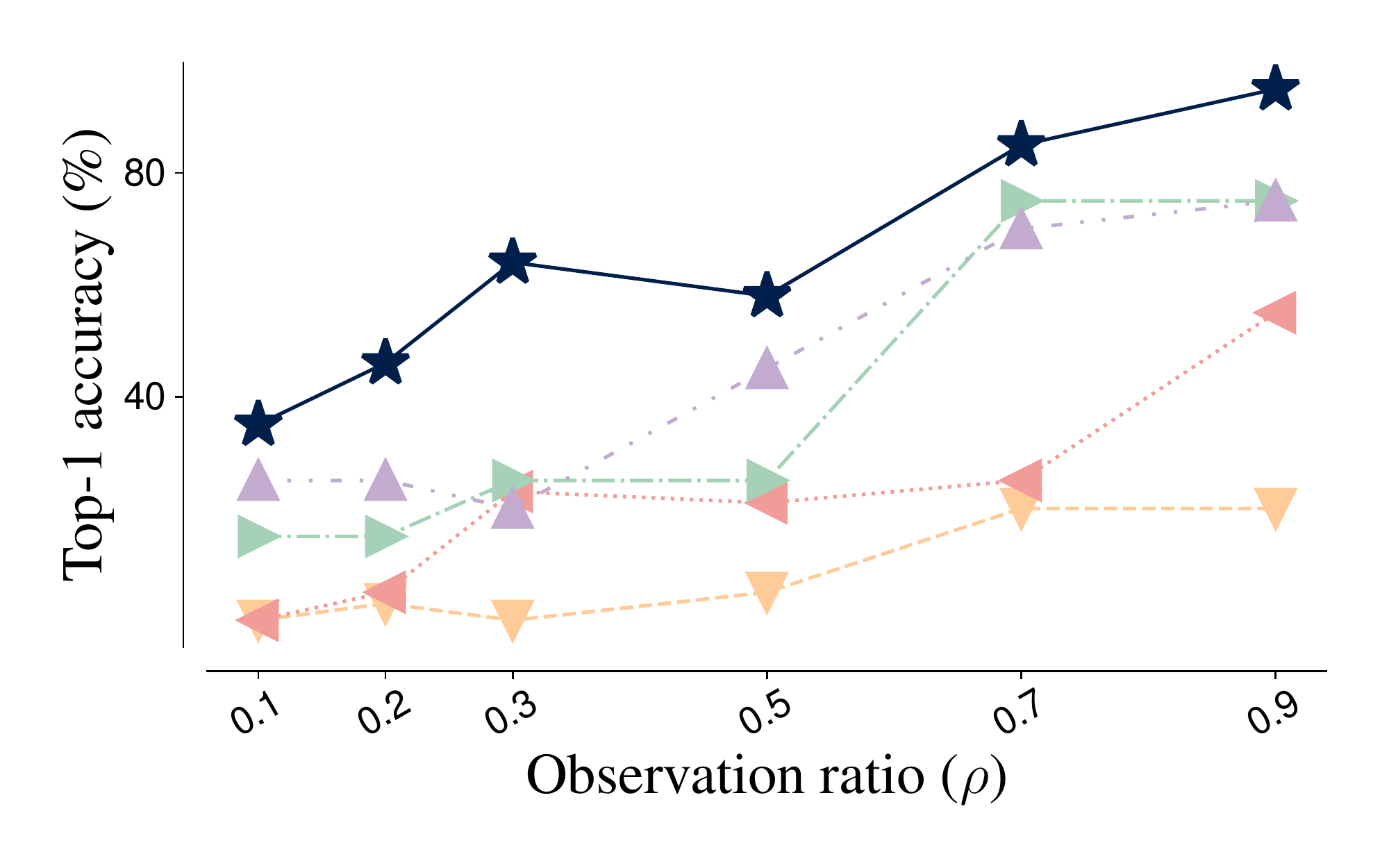}
   \caption{\textit{Poking a stack of something so the stack collapses}}
   \label{fig:poking_something_collapses}
\end{subfigure}
\begin{subfigure}[b]{0.47\textwidth}
   \includegraphics[width=\linewidth]{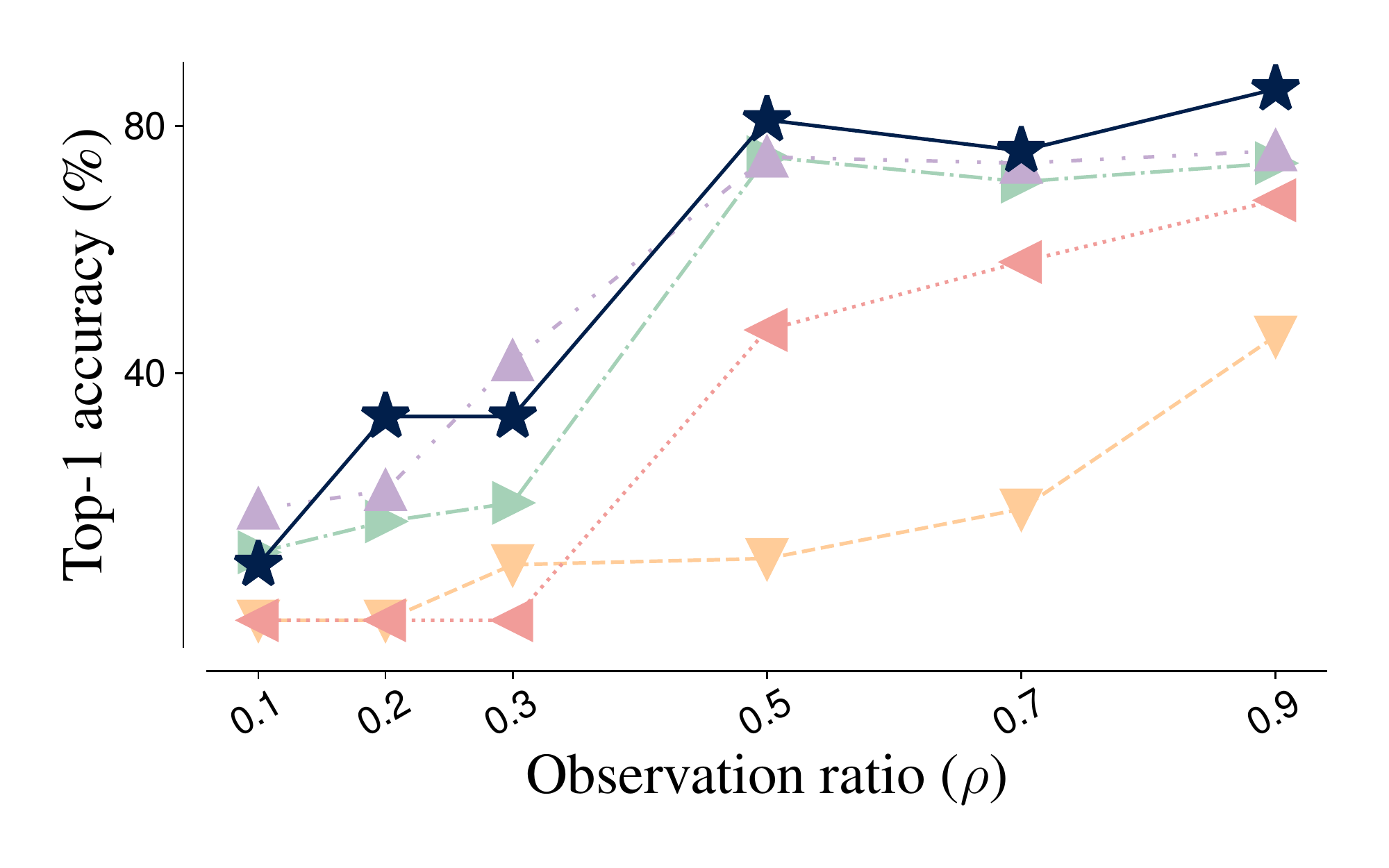}
   \caption{\textit{Poking a stack of something without collapsing}}
    \label{fig:poking_smthng_without_collapsing}
\end{subfigure}
\caption{\textbf{TemPr} \tempriv ~\textbf{SSsub21 tower accuracies across observation ratios for classes} (a) \textit{Closing Something}, (b) \textit{Opening Something}, (c) \textit{Poking a stack of something so the stack collapses} and (d) \textit{Poking a stack of something without collapsing}.}
\label{fig:ssub21_4cases}
\end{figure}

\noindent
\textbf{Scale configurations}. Supplementary to Table \textcolor{red}{1} in the main text, we consider the two top-performing backbones in \Cref{tab:ucf101_scales} and ablate over four scale configurations on UCF-101. 

For both models, and across observation ratios, Tempr~\tempriv ~ outperforms all other scale configurations with the most notable improvements on smaller observation ratios. For $\rho=0.1$ Tempr \tempriv ~ demonstrates a +3.1\% improvement from Tempr \tempri ~ on X3D$_M$ and +3.6\% on MoViNet-A4.

\noindent
\textbf{Top tower predictor per class}. To better understand the performance of individual towers $\mathcal{T}_{i}$, we compare their performance across SSsub21 classes. In \Cref{tab:sssub21_class_towers}, we present the top-performing tower for each class across observation ratios. Overall, we observe that towers trained on larger scales ($\mathcal{T}_3$ \temprivabd ~ and $\mathcal{T}_4$ \temprivabc ~) are better suited for classes that also include long-term dependencies. E.g. classes such as \textit{Poking a stack of something without the stack collapsing}, \textit{Pretending to sprinkle air onto something}, \textit{Showing something behind something}, or \textit{Putting something into something}, require a larger part of the action to be observable to become distinguishable. In contrast, towers for smaller scales, are better suited for  classes such as \textit{Picking something up}, \textit{Closing something, or \textit{Turning something upside down}}, which are distinguishable from only a few frames.

\begin{figure*}[t]
\includegraphics[width=.1\linewidth,trim={0 6cm 41cm 1cm},clip]{figures/class_accuracies_ssub21.pdf}
\includegraphics[width=.85\linewidth,trim={5cm 0 0 0},clip]{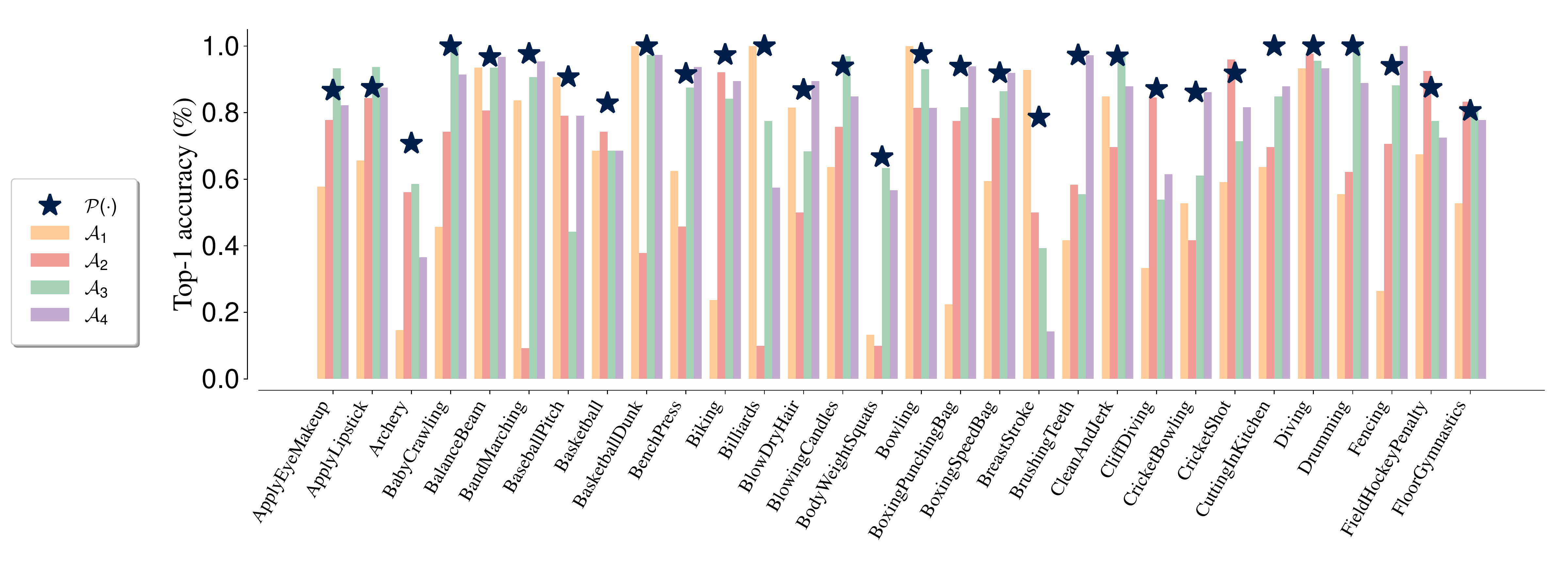}
\caption{\textbf{TemPr} \tempriv ~\textbf{UCF-101 class accuracies for the first 30 classes} over observation ratio $\rho=0.3$.} 
\label{fig:tempr_acc_class_ucf}
\end{figure*}

\noindent
\textbf{SSsub21 class accuracies}. To further determine the performance of tower predictors in \Cref{tab:sssub21_class_towers}, we show in \Cref{fig:tempr_acc_class_sub21} the per-class accuracies of all towers for $\rho=0.3$. Overall, because features are more motion-based compared to UCF-101, coarser scales perform better. Considering the \textit{Putting something on the edge of something so it is not supported and falls down} class, the object will typically fall down only at the end of the action. Therefore, such information is better captured by the coarser scales. Similarly, for \textit{Pretending to sprinkle air onto something}, \textit{pretending} can only be captured over a longer temporal scale. Fine scales perform more favorably for shorter actions such as \textit{Closing something}, \textit{Picking something up}, and \textit{Turning something upside down}. For the majority of these classes, informative motions only last a few frames and are thus better addressed by finer scales. Additionally, in \Cref{fig:ssub21_4cases} we observe that TemPr \tempriv ~ relies more on coarser scales to capture the differences between visually similar classes. Considering the pairs \textit{Closing something} from \Cref{fig:closing_something} and \textit{Opening something} from \Cref{fig:opening_something}, as well as \textit{Poking a stack of something so the stack collapses} from \Cref{fig:poking_something_collapses} and \textit{Poking a stack of something without the stack collapsing} in \Cref{fig:poking_smthng_without_collapsing}, there is a stronger reliance to $\mathcal{T}_4$ \temprivabc ~ and $\mathcal{T}_3$ \temprivabd ~ , with $\mathcal{T}_2$ \temprivacd ~ only performing better for specific $\rho$.  

\begin{table}[t]
\centering
\caption{Tower acc. UCF101.}
\label{tab:ucf_tower_acc}
\resizebox{.9\textwidth}{!}{%
\begin{tabular}{c|cccccc}
\hline
\multicolumn{1}{c|}{\multirow{2}{*}{$\mathcal{T}/\mathcal{E}$}}&
\multicolumn{6}{c}{$\rho$}\tstrut\\
&
0.1 &
0.2 &
0.3 &
0.5 &
0.7 &
0.9\bstrut \\
\hline
$\mathcal{T}_4$~\temprivabc~ & 
78.5& 
82.3& 
86.3& 
84.1& 
89.3& 
87.7 \tstrut \\
$\mathcal{E}(\cdot)$ & 
\textbf{84.3}& 
\textbf{90.2}& 
\textbf{90.4}& 
\textbf{91.2}& 
\textbf{92.1}& 
\textbf{92.4}\\
\end{tabular}
}
\end{table}

\begin{table}[t]
\centering
\caption{Tower acc. SSsub21.}
\label{tab:sssub21_tower_acc}
\resizebox{.9\textwidth}{!}{%
\begin{tabular}{c|cccccc}
\hline
\multicolumn{1}{c|}{\multirow{2}{*}{{$\mathcal{T}/\mathcal{E}$}}}&
\multicolumn{6}{c}{$\rho$}\tstrut\\
&
0.1 &
0.2 &
0.3 &
0.5 &
0.7 &
0.9\bstrut \\
\hline
$\mathcal{T}_4$~\temprivabc~ & 
26.0& 
31.6& 
34.1& 
36.9& 
40.6& 
45.2 \tstrut \\
$\mathcal{E}(\cdot)$ &
\textbf{28.4}& 
\textbf{34.8}& 
\textbf{37.9}& 
\textbf{41.3}& 
\textbf{45.8}& 
\textbf{48.6} \\
\end{tabular}
}
\end{table}

\noindent
\textbf{UCF-101 class accuracies}. In \Cref{fig:tempr_acc_class_ucf}, we present accuracies for the first 30 classes on UCF-101. Overall, the performance of the aggregation function is equivalent to that of the top-performing tower. For the \textit{BreastStroke} class, the finer scale $\mathcal{T}_{1}$ \temprivbcd ~ outperforms other tower predictors. This is also the case for the \textit{Billiards} class which shows a similar trend with $\mathcal{T}_{1}$ \temprivbcd ~ achieving the best performance. We believe the high accuracy over the fine scales of both \textit{BreastStroke} and \textit{Billiards} classes, is due to their unique appearance and motion features. Thus, for only a small portion of the video, the ongoing action can be correctly predicted. 

\noindent
\textbf{Tower and aggregation function accuracies}. Motivated by class accuracy trends observed in \Cref{fig:tempr_acc_class_ucf} and \Cref{fig:tempr_acc_class_sub21} for UCF-101 and SSsub21, we compare the performance of the final attention tower $\mathcal{T}_{4}$ \temprivabc~ to that of the $\mathcal{E}(\cdot)$ aggregator from TemPr~\tempriv~. Results for UCF-101 are presented in \Cref{tab:ucf_tower_acc} and for SSsub21 in \Cref{tab:sssub21_tower_acc}. Consistent improvements are observed by the predictor ensemble compared to the predictions made from individual towers.

\begin{table}[t]
\vspace{-3pt}
\caption{\textbf{Tower designs}.}
\label{tab:ucf101_towers}
\centering
\vspace{7.5pt}
\resizebox{.45\linewidth}{!}{%
\begin{tabular}{l | c c}
\hline
\multicolumn{1}{c|}{\multirow{1}{*}{Tower}} &
\multicolumn{2}{c}{$\rho$} \tstrut \\[.2em]
\multicolumn{1}{c|}{\multirow{1}{*}{design}} &
\multicolumn{1}{c}{0.2} &
\multicolumn{1}{c}{0.4} \bstrut \\
\hline
MLP $\times 4 \,$ &
72.4 & 
81.1 \tstrut \\[.1em]
MLP $\times 8 \,$ &
73.1 &
81.3 \bstrut \\[.1em]
\hline
\textbf{(ours)} &
90.2 &
90.9 \tstrut \\
\end{tabular}
}
\end{table}

\begin{figure}[t]
\begin{center}
\vspace{-1em}
\includegraphics[width=.9\linewidth]{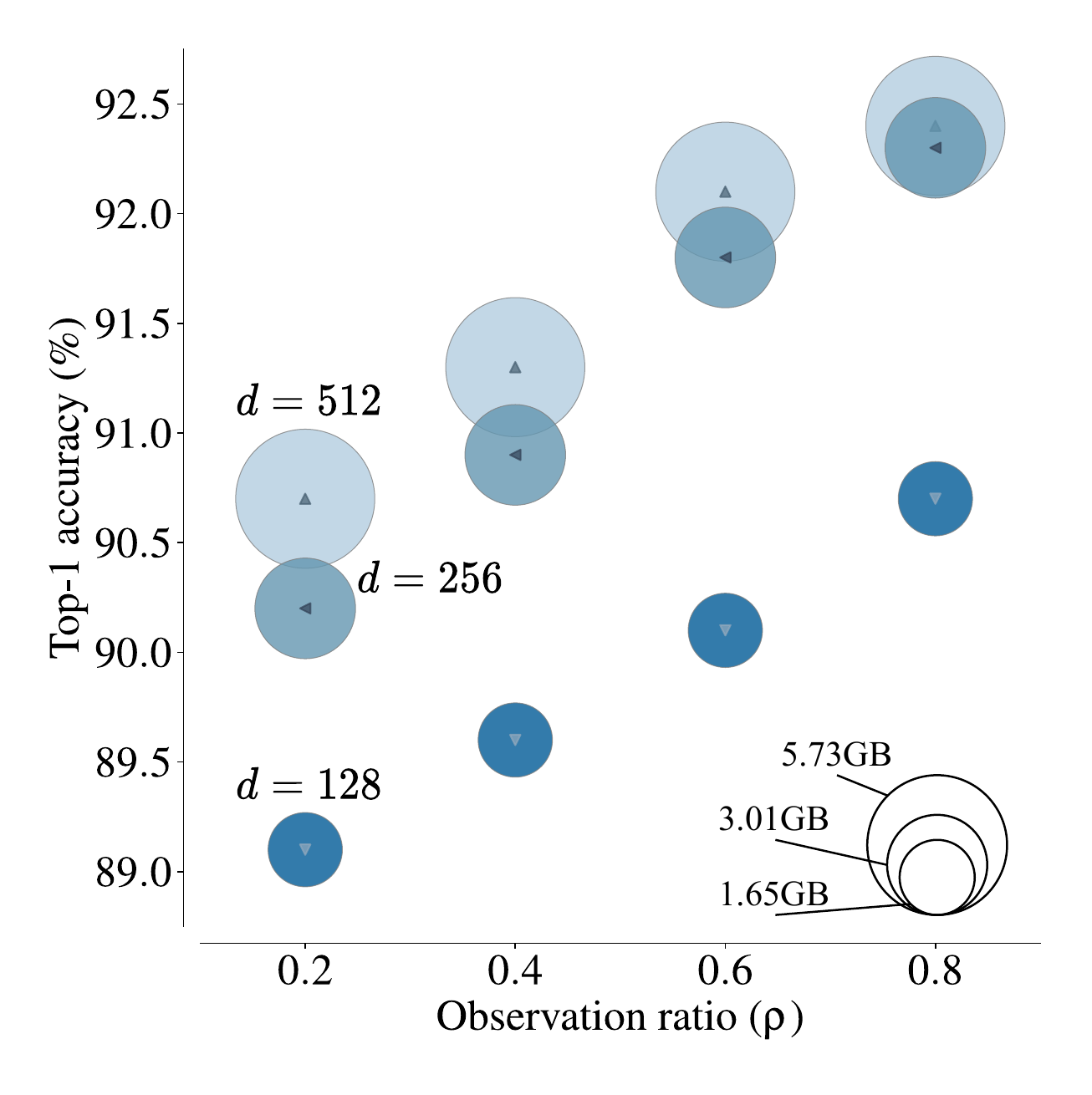}
\end{center}
\vspace{-2em}
\caption{\textbf{Bottleneck size} ($d$) for latent array ($\mathbf{u}$).}
\label{fig:ucf101_latentsize}
\end{figure}

\begin{table}[t]
\caption{\textbf{Bottleneck size comparison} based on latent array ($\mathbf{u}$) index dimension ($d$) used by the cross-attention blocks.}
\label{tab:ucf101_latent}
\centering
\resizebox{\linewidth}{!}{%
\begin{tabular}{c| c | l l l l }
\multicolumn{1}{c|}{\multirow{2}{*}{\shortstack{$d$}}} &
\multicolumn{1}{c|}{Mem.} &
\multicolumn{4}{c}{Observation ratios ($\rho$)} \\[.3em]
&
\multicolumn{1}{c|}{(GB)} &
0.2 &
0.4 &
0.6 &
0.8 \bstrut \\[.3em]
\hline
128 & 
1.65 &
89.1 \textcolor{cadmiumred}{(-1.1)} &
89.6 \textcolor{cadmiumred}{(-1.3)} &
90.1 \textcolor{cadmiumred}{(-1.7)} &
90.7 \textcolor{cadmiumred}{(-2.3)} \tstrut \\[.1em]
256 &
3.01 &
90.2 &
90.9 &
91.8 &
92.3 \\[.1em]
512 & 
5.74 &
\textbf{90.7} \textcolor{applegreen}{(+0.3)} &
\textbf{91.3} \textcolor{applegreen}{(+0.4)} &
\textbf{92.1} \textcolor{applegreen}{(+0.3)} &
\textbf{92.4} \textcolor{applegreen}{(+0.1)} \\[.1em]
\end{tabular}
}
\end{table}

\section{Further ablations}

As with the ablation results in Section \textcolor{red}{4.3} of the main text, we use TemPr \tempriv ~ with ResNet-18 backbone on UCF-101 for all experiments in this section.

%tower designs
\noindent
\textbf{Cross-attention layer replacements}. We include tower ablations in Table~\ref{tab:ucf101_towers} with $\times 4/8$ MLP layers to assess if the improvements are indeed due to our design. A notable drop is observed with the replacement of the attention towers. 

\noindent
 \textbf{Latent array} $\mathbf{u}$ \textbf{size}: In \Cref{fig:ucf101_latentsize} we present performance results on UCF-101 given different latent array $\mathbf{u}$ sizes $d$. Size $d=256$ is shown to be the most cost-effective size as improvements over $d=128$ range between (1.1-2.3)\% while requiring $\sim \!\! 50\%$ less memory than $d=512$. We additionally detail numerically these individual performances in \Cref{tab:ucf101_latent}. In terms of memory, $d=128$ requires $1.36$GB less than $d=256$, while $d=512$ uses $2.73$GB more.

\begin{table}[t]
\caption{\textbf{Number of self attention blocks} (L)} 
\label{tab:ucf101_ablate_selfattention}
\centering
\vspace{0.1em}
\resizebox{\textwidth}{!}{%
\begin{tabular}{c| c c c c c| c c c c }
\multicolumn{1}{c|}{\multirow{2}{*}{L}} &
\multicolumn{2}{c}{Latency (secs)}&
\multicolumn{1}{c}{\multirow{2}{*}{\shortstack{Pars \\ (M)}}} &
\multicolumn{1}{c}{\multirow{2}{*}{\shortstack{FLOPs \\ (G)}}} &
\multicolumn{1}{c|}{\multirow{2}{*}{\shortstack{Mem.\\ (GB)}}} &
\multicolumn{4}{c}{$\rho$} \\
&
I ($\downarrow$) & B ($\uparrow$) &
&
& 
&
0.2 &
0.4 &
0.6 &
0.8 \bstrut \\
\hline
1 &
0.31 & 1.07 &
20.3 &
1.29 &
2.74 &
70.9 &
74.8 &
80.4 &
86.2 \tstrut \\
2 &
0.31 & 1.09 &
20.6 &
1.32 &
2.78 &
77.2 &
76.3 &
82.8 &
86.7 \\
4 &
0.32 & 1.12  &
21.5 &
1.37 &
2.85 &
83.4 &
84.9 &
85.1 &
87.4 \\
6 &
0.32 & 1.16 &
22.2 &
1.42 &
2.93 &
88.7 &
89.5 &
89.8 &
90.1 \\
8 &
0.34 & 1.27 &
23.0 &
1.47 &
3.01 &
\textbf{90.2} &
\textbf{90.9} &
\textbf{91.8} &
\textbf{92.3} \\
\end{tabular}
}
\vspace{-1em}
\end{table}

\begin{table}[t]
\caption{\textbf{Ablation on aggregation function.}}
\begin{subtable}[t]{0.4\textwidth}
\vspace{-3pt}
\subcaption{SSsub21.}
\label{tab:sssub21_ablate_ensemble}
\centering
\resizebox{\textwidth}{!}{%
\vspace{7.5pt}
\begin{tabular}{l| c c }
\hline
\multicolumn{1}{c|}{\multirow{2}{*}{Aggregation}} &
\multicolumn{2}{c}{$\rho$} \tstrut \\[.3em]
&
0.2 &
0.5 \bstrut \\
\hline
avg & 
32.3 &
38.6 \tstrut \\[.2em]
softmax &
31.4 &
36.8 \\[.2em]
ICW &
32.4 &
38.8 \bstrut \\
\hline
\noindent \textbf{adapt.} ($\mathcal{E}(\cdot)$) &
\textbf{34.8} &
\textbf{41.3} \tstrut \\
\end{tabular}
}
\end{subtable}
\hfill 
\begin{subtable}[t]{0.58\textwidth}
\vspace{-3pt}
\subcaption{EK-100.}
\label{tab:ek100_ablate_ensemble}
\centering
\resizebox{\textwidth}{!}{%
\vspace{7.5pt}
\begin{tabular}{l| ccc | ccc }
\hline
\multicolumn{1}{c|}{\multirow{3}{*}{Aggregation}} &
\multicolumn{6}{c}{$\rho$} \tstrut \\[.3em]
&
\multicolumn{3}{c|}{0.2} &
\multicolumn{3}{c}{0.5} \bstrut \\
& V & N & A & V & N & A\\
\hline
avg & 
21.5 & 23.9 & 8.8 &
51.3 & 42.2 & 27.5 \tstrut \\[.2em]
softmax &
19.4 & 23.1 & 8.3 &
50.7 & 41.4 & 24.6 \bstrut \\
\hline
\noindent \textbf{adapt.} $\mathcal{E}(\cdot)$ &
\textbf{22.5} & \textbf{25.5} & \textbf{9.8} &
\textbf{54.2} & \textbf{43.4} & \textbf{28.9}  \tstrut \\
\end{tabular}
}
\end{subtable}
\end{table}

 % Self attention blocks
 \noindent
\textbf{Number of self attention blocks}. \Cref{tab:ucf101_ablate_selfattention} demonstrates the impact of the Self MAB number on the accuracy. Increasing the number of self-attention blocks improves accuracy mostly in small observation ratios, while marginally
increasing the complexity and memory requirements. We, therefore, adopt $L=8$ for our model. 

\begin{table}[t]
\caption{\textbf{Ablating contributions} with individual and combined replacement.}
\label{tab:multi_setting_ablate}
\centering
\resizebox{1\linewidth}{!}{%
\begin{tabular}{c c c | c c c c}
\multicolumn{3}{c|}{replacement(s)} & \multicolumn{4}{c}{\multirow{3}{*}{Obs. ratio ($\rho$)}} \\[.1em]
I. & II. & III. & & & & \\[.1em]
$\mathbf{s}_{1,...,n}$ &
$f(\widehat{\mathbf{z}}_{i})$ &
$\mathcal{E}(\mathbf{y}_{1,...,n}))$ &
& & & \\
$\downarrow$ & 
$\downarrow$ & 
$\downarrow$ & 
\multirow{2}{*}{0.2} &
\multirow{2}{*}{0.4} &
\multirow{2}{*}{0.6} &
\multirow{2}{*}{0.8} \\
$s_{n}\! \times \! n$&
$f(\mathbf{z}_{i})$ &
$\overline{f(\widehat{\mathbf{z}})}$& 
& & & \bstrut \\[.1em]
\hline
\multicolumn{3}{c}{\cellcolor{blue!10} \textbf{Proposed} } &
\cellcolor{blue!10} \textbf{90.2} & 
\cellcolor{blue!10} \textbf{90.9} & 
\cellcolor{blue!10} \textbf{91.8} & 
\cellcolor{blue!10} \textbf{92.3} \tstrut \bstrut \\[.2em]
\hline
\xmark & & &
86.4 & 
88.3 & 
88.8 & 
89.0 \tstrut \\[.2em]
& \xmark & & 
69.4 & 
73.2 & 
78.6 & 
85.5 \\[.2em]
& & \xmark &
89.5 & 
90.1 & 
90.6 & 
91.2 \bstrut \\[.2em]
\hline
\xmark & \xmark & &
64.3 & 
69.8 & 
75.9 & 
83.4 \tstrut \\[.1em]
& \xmark & \xmark & 
67.4 & 
72.8 & 
77.3 & 
84.7 \\[.1em]
\xmark & & \xmark &
84.2 & 
87.0 & 
87.4 & 
88.3 \bstrut \\[.1em]
\hline
\xmark & \xmark & \xmark &
61.4 & 
67.2 & 
73.5 & 
79.3 \tstrut \\[.1em]
\end{tabular}
}
\end{table}

% Combined experiments
\noindent
\textbf{SSsub21 and EK-100 aggregation functions}. Supplementary to the results in Table~\textcolor{red}{3b} for different aggregation functions on UCF-101, we induce additional ablations for SSsub21 and EK-100 in \Cref{tab:sssub21_ablate_ensemble} and \Cref{tab:ek100_ablate_ensemble} respectively. Across both datasets, our proposed adaptive predictor accumulation $\mathcal{E}(\cdot)$performs favorably compared to other aggregation methods. An average improvement of $+5.4\%$ and $+3.8\%$ is observed for UCF-101 and SSsub21.

% Combined experiments
\noindent
\textbf{Combined ablations}. Motivated by Table~\textcolor{red}{3} in the main paper, we present combined changes in the model configuration based on our contributions. Setting I. replaces the progressive scales with $n$ copies of the observable video, $\mathbf{s}_{1,...,n} \rightarrow \mathbf{s}_{n} \! \times \! n$. In setting II. the class predictions are made from the extracted CNN features without the utilization of the attention towers $f(\widehat{\mathbf{z}}_{i}^{L}) \! \rightarrow \! f(\mathbf{z}_{i})$. For setting III. the predictor aggregation function is replaced by averaging classifier predictions $\mathcal{E}(f(\widehat{\mathbf{z}}_{1,...,n})) \! \rightarrow \! \overline{f(\widehat{\mathbf{z}})}$. On average, a 14.63\% accuracy reduction is observed across ratios when predictions are made directly from CNN features. This drop is further amplified when progressive sampling is not used, demonstrating the importance of both the proposed architecture and sub-sampling approach.

\section{Predictor aggregation $\beta$ values}

Our proposed adaptive predictor aggregation function relies on a combination of the similarity of predictor probability distributions and their confidences. The trainable parameter of the function defined in Eq. \textcolor{red}{7} is $\beta$ which determines the potion of $\underset{eICW}{\mathcal{E}(\cdot)}$ and $\underset{eM}{\mathcal{E}(\cdot)}$ that are used for composing the final aggregated probability distribution.

\begin{figure}
    \centering
    \subfloat[TemPr \temprii]{{\includegraphics[clip,trim={0 0 0 0},width=.32\textwidth]{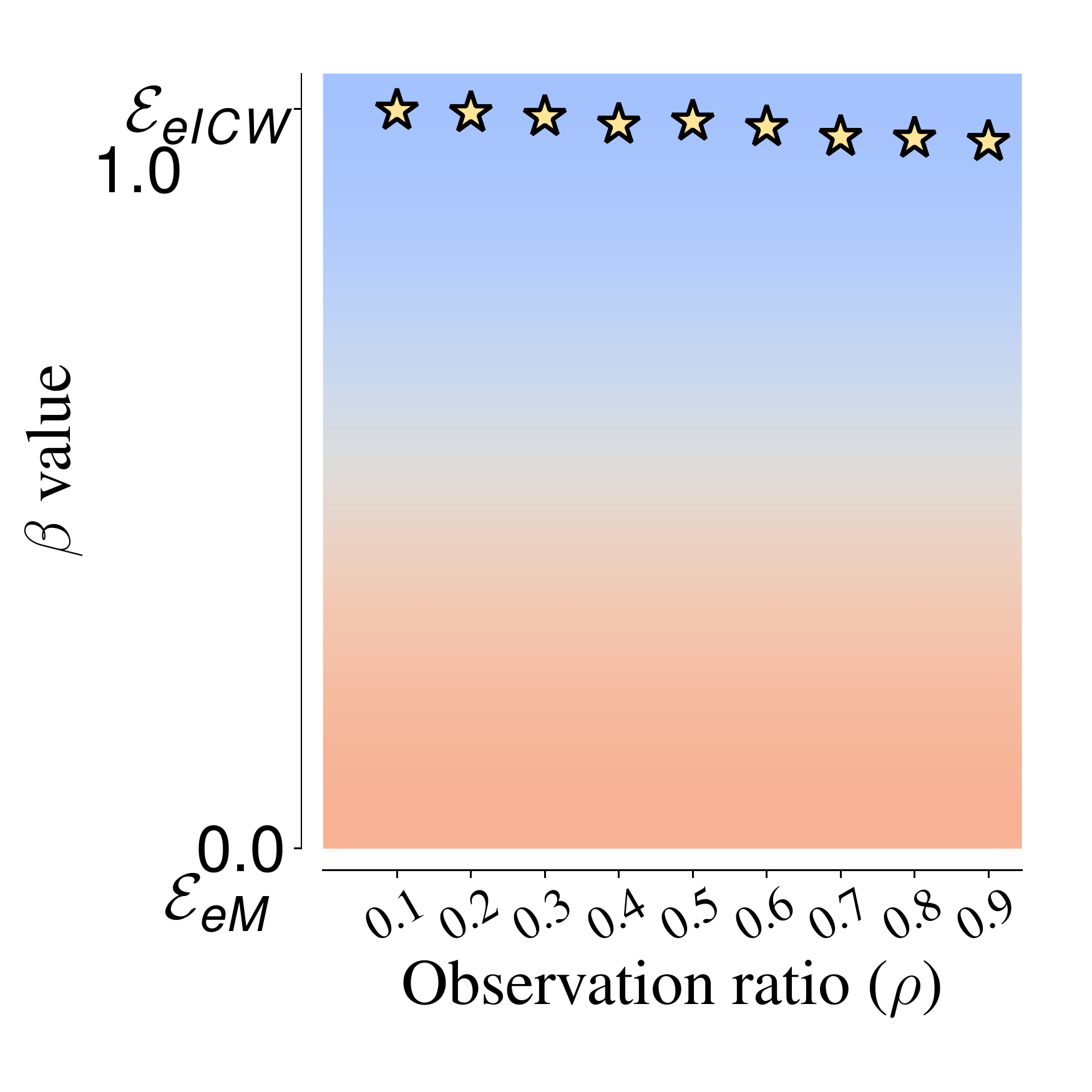}}}
    \subfloat[TemPr \tempriii]{{\includegraphics[clip,trim={0 0 0 0},width=.32\textwidth]{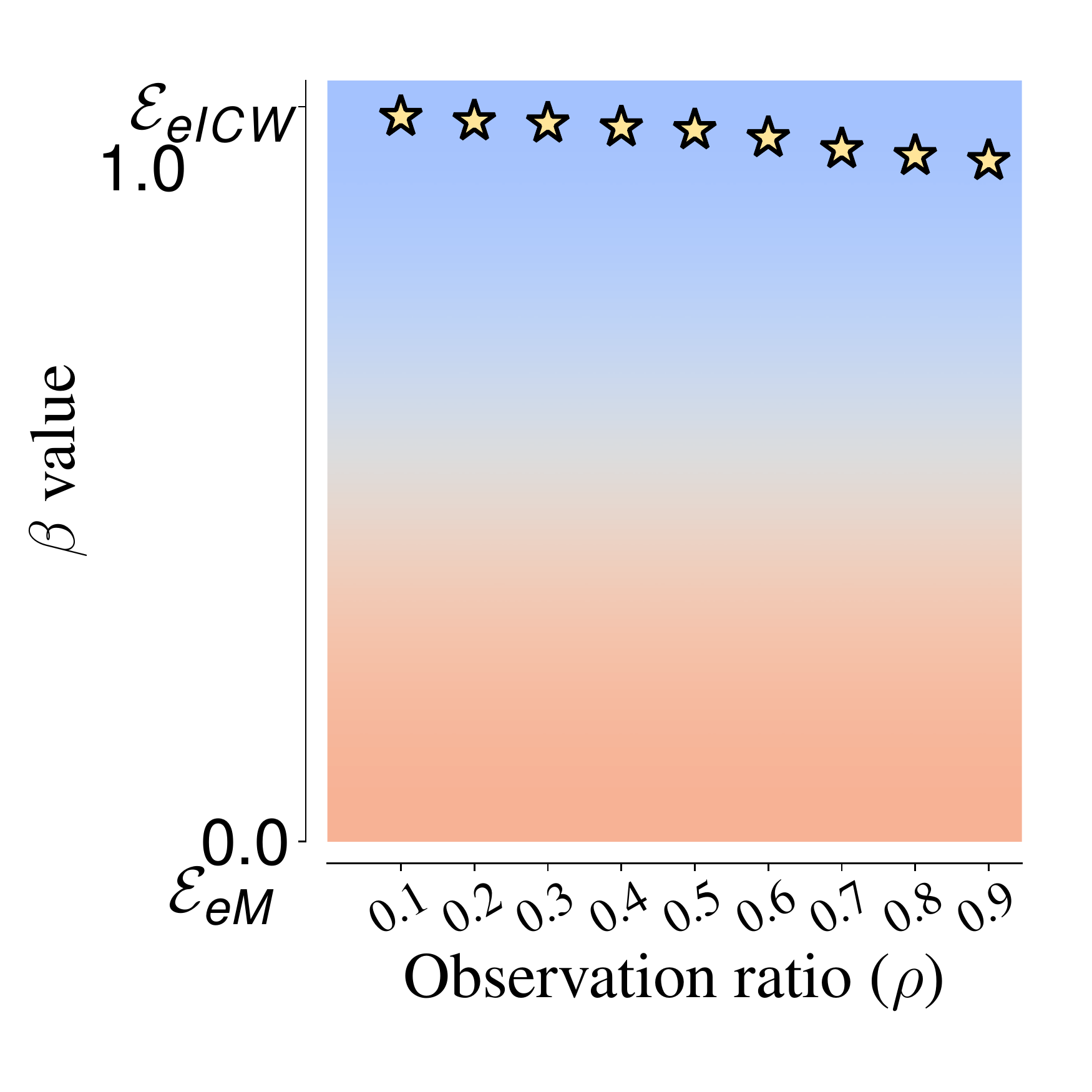}}}
    \subfloat[TemPr \tempriv]{{\includegraphics[clip,trim={0 0 0 0},width=.32\textwidth]{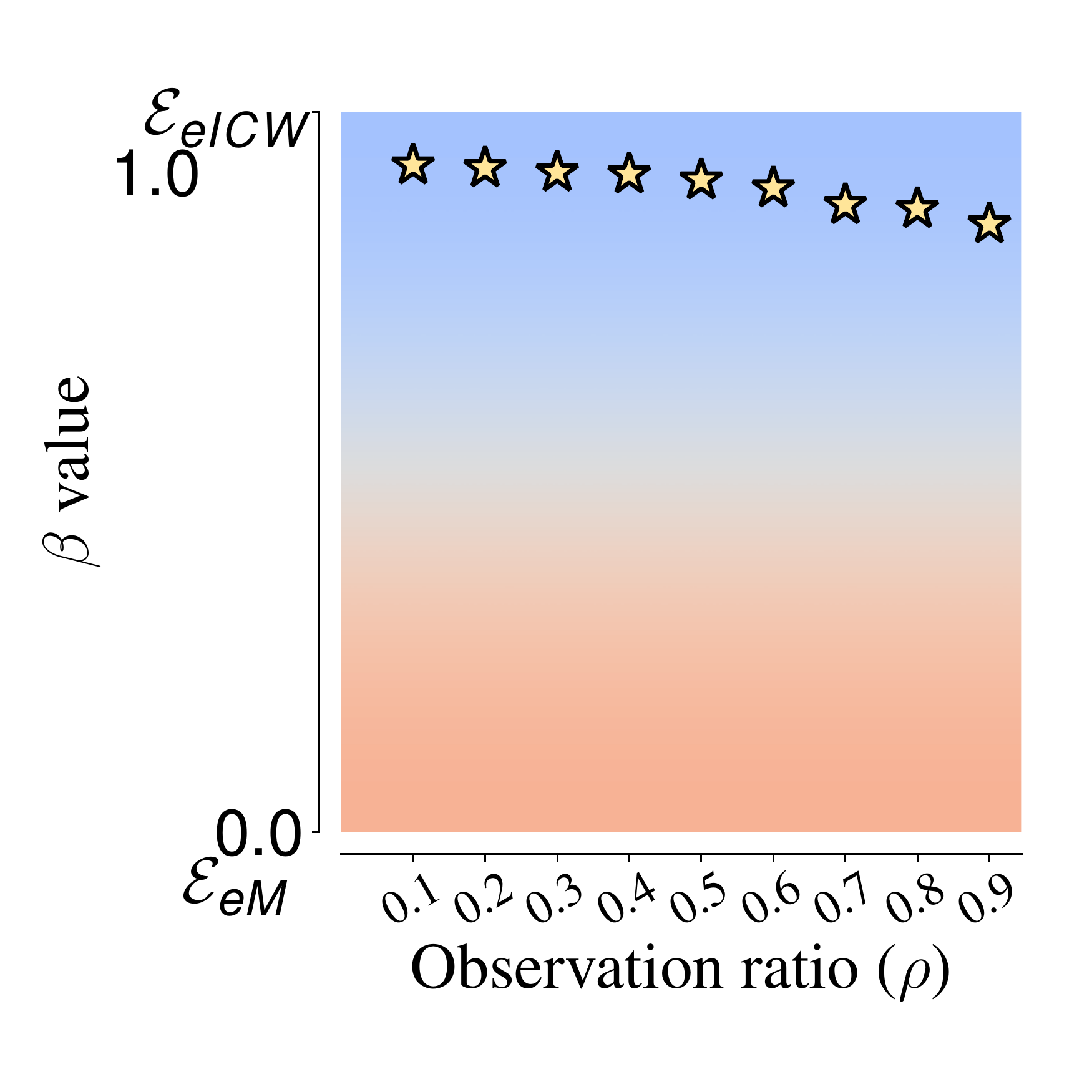}}}
    \caption{\textbf{Post-training $\beta$ values} over obs. ratios on UCF-101.}
    \label{fig:beta_obs_ratio}
\end{figure}

We visualize the values of the $\beta$ parameter, for each TemPr configuration that employs multiple scales (\temprii \ , \tempriii \ and \tempriv \ ) across observation ratios in \Cref{fig:beta_obs_ratio}. We use the  UCF-101 TemPr models with MoViNet-A4. In general, the $\beta$ value remains high within 0.98--0.84 for all observation ratios. A small decrease is observed in larger $\rho$, as independent predictors are exposed to larger portions of the video and can better predict the ongoing action individually. 

\section{Additional Qualitative results over tower predictions}

We have presented and discussed qualitative results over TemPr \tempri \ , \temprii \ , \tempriii \ , \tempriv \ configurations and individual towers $\mathcal{T}_1$ \temprivbcd \ , $\mathcal{T}_2$ \temprivacd \ , $\mathcal{T}_3$ \temprivabd \, $\mathcal{T}_4$ \temprivabc \ in Section~\textcolor{red}{4.3}. Here we provide additional examples in the same format as Figure \textcolor{red}{4}, where predictions differ across TemPr \tempriv \ towers. 

As shown in \Cref{fig:tempr_acc_instances_ucf101}, presented over 2 pages, our proposed progressive scales can benefit feature modeling for a variety of action instances e.g. for the \textit{Lunges} instance, the finer scales ($\mathcal{T}_1$ \temprivbcd \ and $\mathcal{T}_2$ \temprivacd \ ) focus on smaller motions and thus are less influenced by global motion in the video. For \textit{Lunges} and \textit{IceDancing} (form UCF-101), these global motions are similar to those performed for \textit{BodyWeightSquats} and \textit{SalsaSpin}. On the other hand, for the \textit{HighJump} and \textit{SkateBoarding} instances from UCF-101, as well as \textit{hopping} in NTU-RGB and \textit{Pretending to turn something upside down} and \textit{Closing something} in SSsub21, coarse scales are better suited, as motions over larger temporal lengths are more descriptive of the action performed. Failure cases for coarse scales are evident in the chosen examples of \textit{ShavingBeard} from UCF-101, \textit{wipe face} in NTU-RGB, and \textit{turn-off tap} in EPIC-KITCHENS-100, where motions that are descriptive for the class, are performed fast and over shorter temporal durations.

\begin{figure*}[!htbp]
\includegraphics[width=\linewidth]{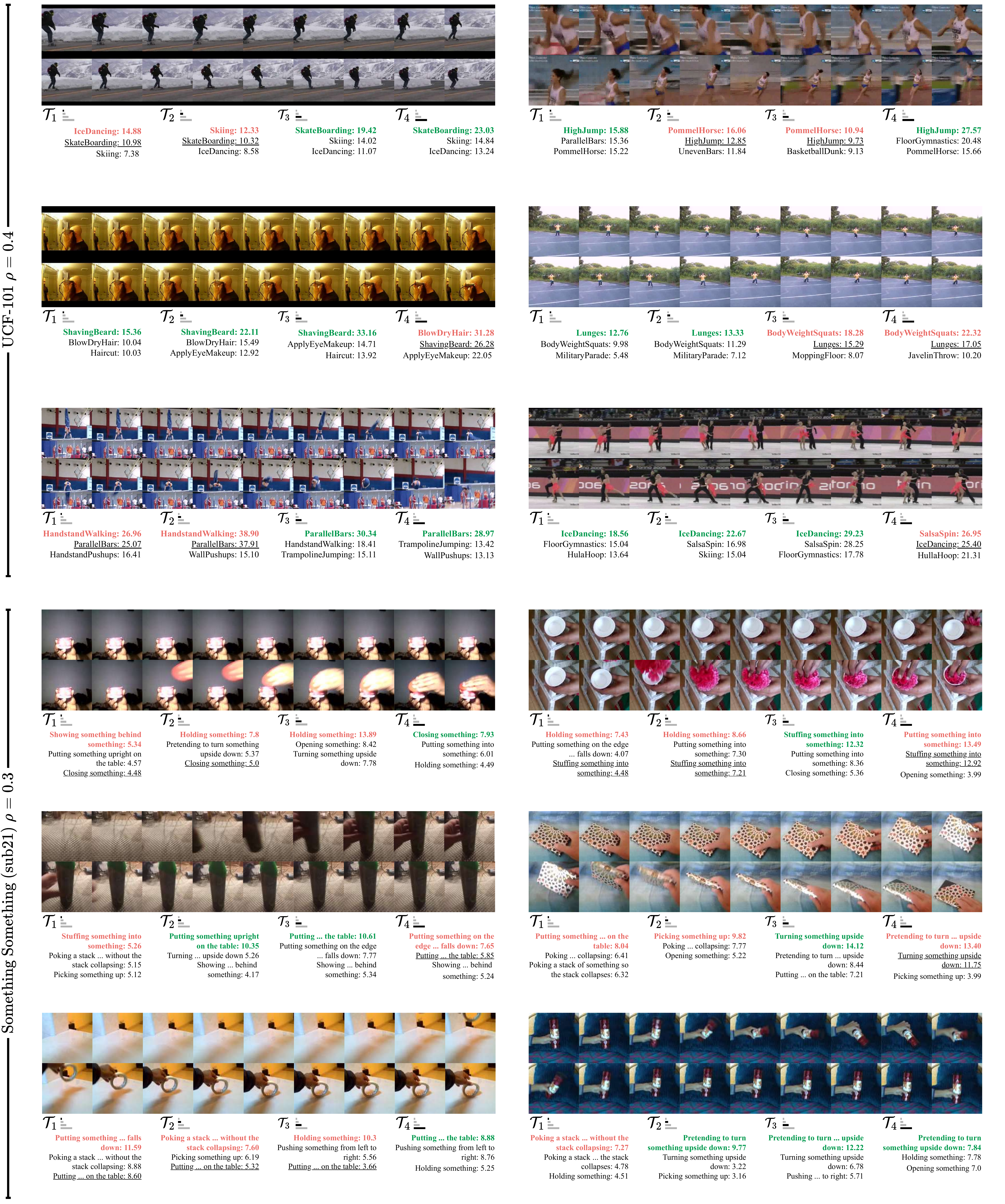}
\caption{\textbf{Instances over UCF-101, SSsub21, NTU-RGB and EK-100}. Top 3 action labels are reported for individual tower predictors $\mathcal{T}_i$ (continues to the next page).} 
\end{figure*}%
\begin{figure*}[!htbp]\ContinuedFloat
\includegraphics[width=\linewidth]{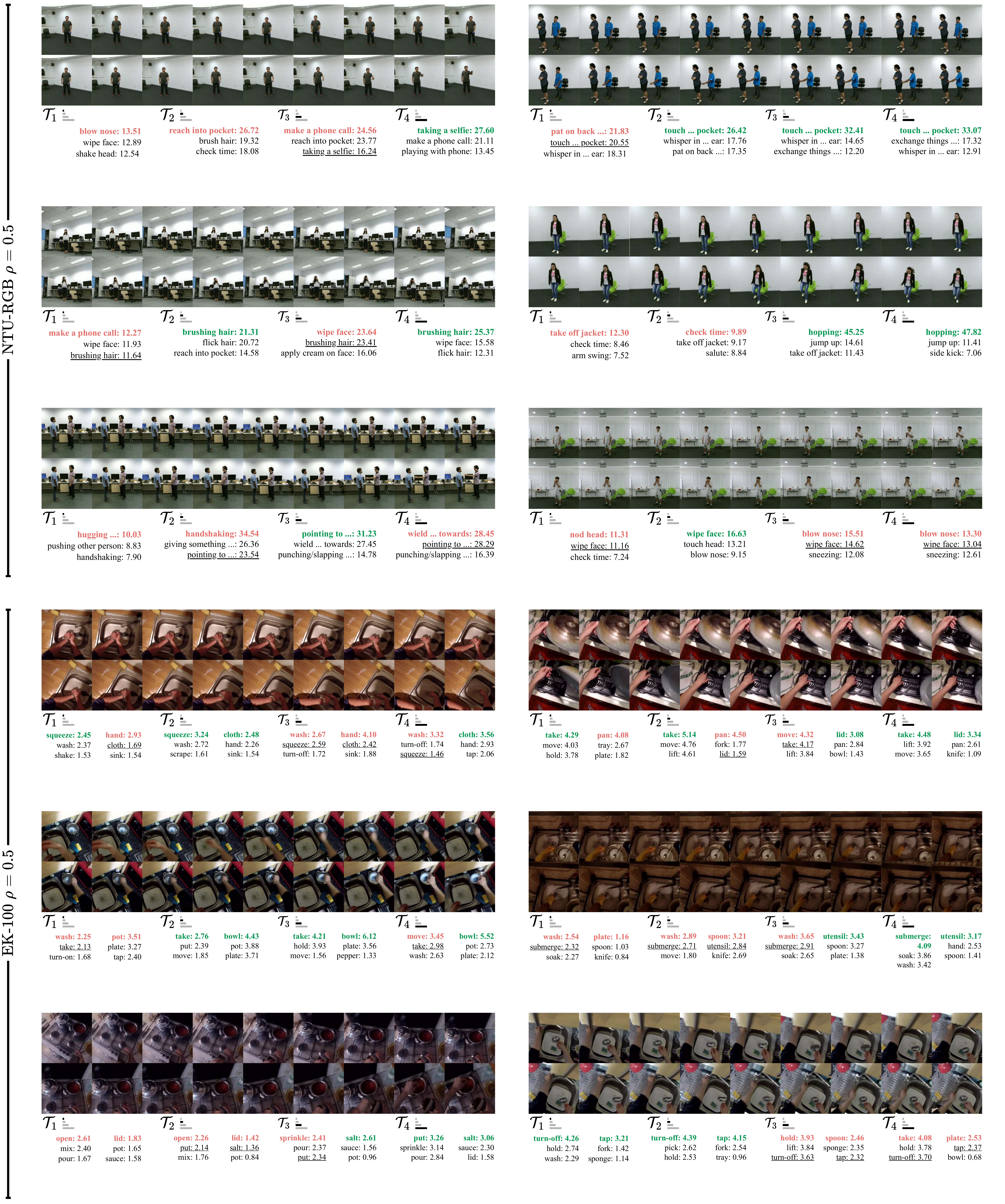}
\caption{\textbf{Instances over UCF-101, SSsub21, NTU-RGB and EK-100}. Top 3 action labels are reported for individual tower predictors ($\mathcal{T}_i$).} 
\label{fig:tempr_acc_instances_ucf101}
\end{figure*}

\end{document}